%% file: acl_latex.tex
\pdfoutput=1

\documentclass[11pt]{article}

\usepackage[]{acl}

\usepackage{times}
\usepackage{latexsym}

\usepackage[T1]{fontenc}

\usepackage[utf8]{inputenc}

\usepackage{microtype}

\usepackage{inconsolata}

\usepackage{algorithm}
\usepackage{algorithmic}
\usepackage{diagbox}
\usepackage{amssymb}
\usepackage{amsmath}
\usepackage{bbm}
\usepackage{url}
\usepackage{adjustbox}
\usepackage{multicol}
\usepackage{multirow}
\usepackage{booktabs}
\usepackage{graphicx}
\usepackage{makecell}
\usepackage{graphicx}
\usepackage{subcaption}
\usepackage{cleveref}
\usepackage{color} 
\usepackage{tabularx}

\title{Reinforcement Learning with Token-level Feedback\\ for Controllable Text Generation}

\author{
Wendi Li\textsuperscript{12}, Wei Wei\textsuperscript{12}\thanks{*Corresponding author.}, Kaihe Xu\textsuperscript{3}, Wenfeng xie\textsuperscript{3}, Dangyang Chen\textsuperscript{3}, Yu Cheng\textsuperscript{4}\\
\textsuperscript{1}Cognitive Computing and Intelligent Information Processing Laboratory,\\
School of Computer Science and Technology, Huazhong University of Science and Technology\\
\textsuperscript{2}Joint Laboratory of HUST and Pingan Property \& Casualty Research (HPL)\\
\textsuperscript{3}Ping An Property \& Casualty Insurance company of China\\
\textsuperscript{4} The Chinese University of Hong Kong\\
\texttt{\{wendili,weiw\}@hust.edu.cn}
}

\begin{document}
\maketitle

\input{sections/abstract}
\input{sections/introduction}

\input{sections/related}
\input{sections/methods}

\input{sections/experiments}
\input{sections/conclusion}

\bibliography{custom}

\newpage
\appendix

\input{sections/appendix}

\end{document}

%% file: sections/abstract.tex
\begin{abstract}

To meet the requirements of real-world applications, it is essential to control generations of large language models (LLMs). 
Prior research has tried to introduce reinforcement learning (RL) into controllable text generation while most existing methods suffer from overfitting issues (finetuning-based methods) or semantic collapse (post-processing methods). 
However, current RL methods are generally guided by coarse-grained (sentence/paragraph-level) feedback, which may lead to suboptimal performance owing to semantic twists or progressions within sentences.
To tackle that, we propose a novel reinforcement learning algorithm named \textbf{TOLE} which formulates \textbf{TO}ken-\textbf{LE}vel rewards for controllable text generation, and employs a "first-quantize-then-noise" paradigm to enhance the robustness of the RL algorithm.   
Furthermore, TOLE can be flexibly extended to multiple constraints with little computational expense. 
Experimental results show that our algorithm can achieve superior performance on both single-attribute and multi-attribute control tasks. We have released our codes at \url{https://github.com/WindyLee0822/CTG}.
\end{abstract}

%% file: sections/introduction.tex
\section{introduction}
\label{sec:introduction}
Large autoregressive language models (LLMs) trained on extensive corpus can generate high-quality texts. However, to satisfy real-world applications, making the generation more controllable is urgent. It is desired to enhance specific attributes of generated texts for practical needs (e.g. positive sentiment for psychological escort, formality for academic writing) \citep{apply-1,apply-2,pretrain} and reduce intrinsic defects of pre-trained language models (e.g. toxicity, repetition) \citep{toxic-1,toxic-2}. 

Retraining models \cite{cocon,ctrl} are subject to computational overheads as the parameter scales become huge.
Post-processing methods \cite{gedi,fudge,dexpert} leverage small-scale discriminators to bias token distribution, which often leads to low text quality.
Some methods \cite{discup,tailor,prompt_gating} adopt parameter-efficient training strategy e.g. prefix-tuning, but they are susceptible to undesired attributes in the supervised corpus.
Recent research \cite{diffusionlm,distlens,normal_flow} introduces other algorithm backbones e.g. diffusion models, normalized flow, but they generally cost more computational expenses during trainig, and have a longer inference time, thus hard to deploy in real applications.

There is some research \cite{distributional,quark} introducing reinforcement learning (RL) into controllable text generation (CTG) tasks. 
RL paradigms can relieve the above problems, which alleviate the overfitting issue by training on self-generated sentences, and can integrate parameter-efficient strategies with canonical LLM backbones.
However, RL-based methods generally update language models with sentence-level (or paragraph-level) rewards, leading to suboptimal performance and slow convergence. 
The coarse-grained rewards cannot provide clear guidance, since semantic in the sentence often transits with twists or progression. 
Moreover, different parts of the sentence may contribute to different attributes.
Therefore, RL methods with coarse-grained feedback generally require considerable training steps to converge.

Our objective is to granularize the coarse-grained feedback to provide more precise guidance for LLMs. 
In this paper, we propose a novel reinforcement learning algorithm with \textbf{TO}ken-\textbf{LE}vel guidance named \textbf{TOLE}.
We first provide an alternative perspective of Bayesian Factorization, which inspires us to formulate the token-level rewards as the probability shifts of attribute classifiers.
To enhance the robustness of TOLE, we propose an exploration framework with "First quantize, then noise" procedure. 
Moreover, TOLE can be extended to multi-attribute scenarios with few computational overheads. 
We conduct two experiments on single-attribute: sentiment control and detoxification. We also evaluate TOLE on multi-attribute scenarios with two settings. 
TOLE achieves superior performance compared with a wide range of baselines.

%% file: sections/related.tex
\section{Related Works}

\textbf{Controllable Text Generation.} Most previous works on controllable text generation (CTG) are based on the auto-regressive framework, which can be categorized into retraining \cite{ctrl,cocon}, finetuning \cite{prompt_gating,tailor,discup}, and post-processing \cite{gedi,dexpert,fudge}. Retraining and traditional finetuning methods are of low efficiency since the parameter scale of LMs is surging and the overfitting issue is severe. Post-processing methods regulate the next-token distribution with supplementary modules, mostly an attribute discriminator, but often cause syntax interruption and make language models lose insights. \citet{quark} integrate RL algorithms into CTG but use coarse-grained feedback to guide the LLMs.

\textbf{Multi-aspect controllable text generation.}
Along with single-aspect controlling, most research on multi-aspect controllable text generation can also categorized into finetuning and post-processing. Some post-processing research \cite{blend, mucoco} in MCTG combines multiple attribute discriminators to aggregate the controllability. However, they also inherit drawbacks of post-processing methods due to direct distribution regulations.
Finetuning-based research tries to connect several single controllers, e.g. connectors to combine multiple plugins \cite{tailor}, latent variables to represent the unsupervised aspects \cite{contrastive}, direct combination of prompts \cite{prompt_gating}, the boundary exploration of intersected subspaces \cite{distlens, normal_flow}. 
To the best of our knowledge, we are the first to explore how to extend single-attribute reinforcement learning algorithms to the MTCG scenario. 

\textbf{Token-level guidance for Reinforcement Learning.}
There is a series of research \citep{decision,rl_transformer1,rl_transformer2,rl_transformer3} incorporating RL techniques into the transformer structure, trying to deconstruct the coarse-grained reward into the token level for sequential modeling. However, they are hard to extend to practical applications since their specialized token settings are not in line with current LLMs.
Concurrent with our research, some research \cite{fine-grained-rlhf,token-level-rlhf} on LLM alignments tries to handle the problem of coarse-grained feedback. RLHF (reinforcement learning from human feedback) algorithms of the LLM alignment generally require a large-scale reward model, which should be trained on datasets formatted as pairwise sentences with the same prefix.
However, such data is unavailable when confronted with a wide variety of attribute requirements.
Therefore, exploring a novel reinforcement learning algorithm with token-level feedback is significant for controllable text generation.  


%% file: sections/methods.tex
\begin{figure*}
    \centering
    \includegraphics[width=\linewidth]{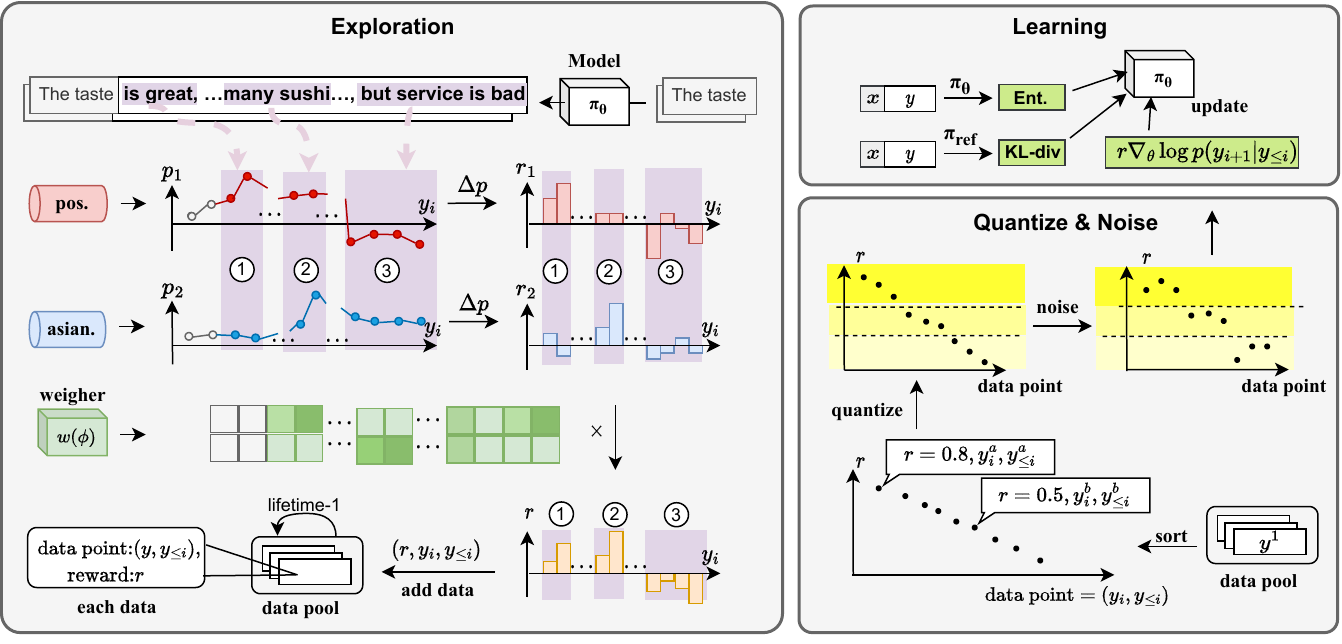}
    \caption{Overall Framework of our algorithm.}
    \label{fig:enter-label}
\end{figure*}

\section{Approach}

We will first establish the notation, provide some background on existing RL methods in controllable text generation and model alignment, and offer an overview of our algorithm.

\subsection{Preliminaries}
\label{sec:RL_formulation}
\textbf{Notations.} A standard Markov Decision Process (MDP) can be denoted as $(\mathcal{S},\mathcal{A},\mathcal{T},r )$. At each step, an action $a \in \mathcal{A}$ is made based on the current state $s\in \mathcal{S}$. Then the state will be transited to $s'$ with the possibility $\mathcal{T}(s'|s,a)$. A function $r: \mathcal{S}\times\mathcal{A} \rightarrow \mathbb{R}$ defines the returned reward based on the states and actions. The strategy is decided by a policy model $\pi(\cdot|s)$, which is a predicted distribution over actions based on state $s$.
To transfer to text generation scenarios, the state can be defined as the partially generated sentence $y_{\leq i-1}=(y_1,y_2,\dots,y_{i-1})$, and the action is the next token $y_i \in \mathcal{V}$ where the vocabulary $\mathcal{V}$ is the action space. The transition dynamic $\mathcal{T}(\cdot|s,a)$ is deterministic since each state-action pair $(y_{\leq i-1},y_i)$ leads to a unique state $y_{\leq i}$. 


\textbf{Prior RL-based methods.} In previous RL-based methods of controllable text generation, rewards are derived from $\mathcal{P}(c|y)$, which denotes the possibility that the sentence $y$ satisfy the attribute $c$. $\mathcal{P}(c|y)$ can be obtained by corresponding attribute classifiers. 
Since prior research only concentrates on sentence-level feedback, which can be regarded as $r(y_1,y_{\leq 0})=r(y_2,y_{\leq 1})=\dots=r(y_{n+1},y_{\leq n})=f(\mathcal{P}(c|y))$. This equality means that sentence-level feedback treats each action $y_i$ in the MDP process of $y$ equally, which can only provide rough guidance for models.

\textbf{Bayesian Factorization in Prior research.} The objective of controllable text generation is to let LLMs approach $\mathcal{P}(y|c)$ where $c$ is a target attribute. Granularize to the token-level, prior post-processing methods generally factorize this term by the Bayesian formula as follows, 
\begin{equation}
    \mathcal{P}(y_{\leq i}|c) \propto \mathcal{P}(c|y_{\leq i})\mathcal{P}(y_i|y_{\leq i-1}).
\end{equation} 
With this formula, post-processing methods can achieve $\mathcal{P}(y|c)$ by regulating the token distribution $\mathcal{P}(y_i|y_{\leq i-1})$ with an attribute classifier which approximates $\mathcal{P}(c|y_{\leq i})$.



\subsection{Token-level Rewards}

We first provide an alternative perspective of Bayesian factorization to show that the probability shift of attribute classifiers plays an important role in controlling the generations.
The Bayesian factorization can be rewritten as:
\begin{equation}
    \mathcal{P}(y_i|y_{\leq i-1},c) 
    \propto \frac{\mathcal{P}(c|y_{\leq i})}{\mathcal{P}(c|y_{\leq i-1})}
    \mathcal{P}(y_i|y_{\leq i-1}).
\label{fml:newb}
\end{equation}
See more details in Appendix \ref{apd:bayesian}. In Eq.\ref{fml:newb}, $\frac{\mathcal{P}(c|y_{\leq i})}{\mathcal{P}(c|y_{\leq i-1})}$ is crucial for the next-token probability distribution. Even if $y_{\leq i}$ tends to highly satisfy the condition $c$ when sentence is finished i.e. $\mathcal{P}(c|y_{\leq i})$ is large, action $y_i$ may not play an important role since previous $y_{\leq i-1}$ may already make future generations satisfy $c$ easily i.e. $\mathcal{P}(c|y_{\leq i-1})$ is large. It reveals that what matters is the probability shift between them, which enlightens our reward design. 

The token-level reward function can be formulated as the probability shift before and after the word is generated.
\begin{equation}
    r(y_{i+1},y_{\leq i}) = f(\mathcal{P}(c|y_{\leq i+1}) - \mathcal{P}(c|y_{\leq i})),
\label{fml:token-level reward}
\end{equation}
where $f(\cdot)$ is an activation function for normalization, where we adopt the sigmoid function for implementations.
Theoretically, to approximate $\mathcal{P}(c|y_{\leq i})$, the format of training data should be transformed from the traditional $\{(y,c)\vert y\in \mathcal{Y}\}$ to $\{(y_{\leq i},c)\vert 0\leq i\leq |y|, y\in \mathcal{Y}\}$ as in \citet{fudge}. However, we find using traditional classifiers in our algorithms can achieve on-par performance in experiments compared to specially trained classifiers. We present this comparison in Appendix \ref{apd:further-fudge}.

\subsection{RL Algorithm: First quantize, then noise.} 
The training procedure of our RL algorithm can be separated into initialization, exploration, quantize \& noise, and learning. 

\textbf{Initialization.} First, we initialize a policy LLM $\pi_\theta$, a copy of the policy model as the reference model $\pi_\textrm{ref}$, an attribute scorer $S$. The reference model is frozen during the whole process. We also initialize a data pool $\mathcal{D}=\varnothing$, and prepare a prefix corpus for exploration. 

\textbf{Exploration.} Then, given the prefix $x$, the current policy model can generate subsequent text $y$. 
For each generated token, we calculate the score shift as its reward $r(y_{i+1},y_{\leq i})$, and add $(y_{i+1},y_{\leq i},r)$ to the data pool $\mathcal{D}$. 
To avoid over-training on data explored in earlier episodes, we set a lifetime for each data to indicate the episodes it can still undergo. Once the data is added to $\mathcal{D}$, the lifetime is initialized to $\mathcal{L}$ and subtracts 1 after each training episode. The data is removed from $\mathcal{D}$ when its lifetime drops to 0.

\textbf{Quantize \& Noise}
Learning primitive rewards $r$ can predispose the model to flatter the scoring pattern of attribute classifiers, which may cause diversity decrease. 
Therefore, we propose "First quantize, then noise" to avoid this problem.
First, we quantize the rewards within $\mathcal{D}$, and acquire $q$-quantiles, which divide the reward range into $q$ intervals. Then, we inject noise into each reward while ensuring each reward stays in the original interval. 
Specifically, for a reward $r \in (q_i,q_{i+1})$, we reassign it as 
\begin{equation}
    \hat{r}=q_i + (q_{i+1}-q_i)\epsilon(r-q_i) 
\label{fml:noise_reward}
\end{equation}
where $\epsilon(\cdot)$ is a noise processed with a clip function to satisfy $\epsilon(r) \in (r-1,r+1)$. $\epsilon(r)$ is substituted with Gaussian noise in our implementations.
Through this process, we disrupt the reward order to interfere the fixed scoring patterns of classifiers, while maintaining the relative order between intervals to steer LLMs toward the target attribute.

\textbf{Learning.} 
Through above procedures, we can obtain $\hat{r}$ to provide dense guidance on each token without granularity mismatch or feedback delay.
The minimalist objective of our optimization problem is to maximize the total rewards, $
\max_\theta \mathbb{E}_{y_{t+1}\sim \pi_\theta(\cdot|y_{\leq t})} [ \hat{r}(y_{t+1},y_{\leq t}) ]
$.
We relax the convergence by adding a standard max-entropy gradient, which can help capture diverse behavior modes. We also insert a KL-divergence penalty to keep the model $\pi_\theta$ from deviating too far from the original $\pi_\textrm{ref}$. 
The gradient of each sentence $y$ can be formulated as follows,
\begin{equation}
\begin{aligned}
 &\mathbb{E}_{y_{t+1}\sim \pi_\theta(\cdot|y_{\leq t})} \Big[ \hat{r}(y_{t+1},y_{\leq t}) \nabla_\theta \log \pi_\theta(y_{t+1}|y_{\leq t}) \\
    &\qquad \qquad +\alpha\nabla_\theta  \mathcal{H}(\cdot|y_{\leq t}) + \beta\nabla_\theta  \textsc{KL}(y_{\leq t}) \Big]
\end{aligned}
\label{fml:loss_function}
\end{equation}
where $\alpha,\beta $ are two balancing coefficient, $\mathcal{H}$ is the Shannon entropy of $\pi_\theta(\cdot|y_{\leq t})$, $\textsc{KL}(y_{\leq t})$ is the KL divergence between $\pi_\theta(y_{t+1}|y_{\leq t})$ and $\pi_\textrm{ref}(y_{t+1}|y_{\leq t})$.

We then use the updated model for exploration and repeat the Exploration-Quantize \& Noise-Learning cycle until training achieves the maximum episode number.


\subsection{Extension to Multiple Attributes.}
To consider multiple constraints simultaneously, we should combine multiple reward groups from different scorers. Simple aggregations or averages cannot provide appropriate token-level guidance, since scorers may contradict each other. Moreover, different parts of sentences may address different attributes, so we need to weigh the token's contribution to multiple attributes respectively.
To tackle this, we train a small-scale "weigher"  $W_\phi: \mathbb{R}^d \rightarrow \mathbb{R}^n$ to balance rewards from $n$ scorers, where $d$ is the hidden size of LLMs. 
Given the last-layer hidden states $\mathbf{H}_{t+1} \in \mathbb{R}^{1\times d}$ of $y_{t+1}$ output by LLMs $\pi(y_{\leq t+1})$, the weigher output $\mathbf{W} = W_\phi(\mathbf{H}_{t+1})$ as the weight for $n$ rewards of $y_t$, $\mathbf{R}_{t+1} \in \mathbb{R}^{1\times n}$. 
The weigher does not require a complex model structure. Simple structures can already assist our algorithm to achieve great performance. Hence it does not take significant computational overheads.
In our implementation, the weigher consists of two linear layers with the ReLU function and a output layer with a softmax function. 
The comprehensive reward of action $y_{t+1}$ can be obtained by $r = \mathbf{W}\times \mathbf{R}^T_{t+1}$.

To train the weigher, we formulate the optimization problem as maximizing the integrated reward of a training corpus $y\sim \mathcal{Y}$ that satisfies the multiple attributes,
\begin{equation}
    \max_\phi \mathbb{E}_{y \sim \mathcal{Y}}  \mathbb{E}_{t} W_\phi(\mathbf{H}_{t+1}) \times \mathbf{R}_{t+1}
\end{equation}
where $t\sim \textrm{Uniform}(0,|y|-1)$, a uniform distribution among $\{0,1,\dots,|y|-1\}$. By doing so, the weigher learns which scorer should be paid more attention when considering different tokens within sentences.

%% file: sections/experiments.tex
\section{Experiments}
\label{sec:exp}

\begin{table*}[t!]
    \centering\footnotesize
    \tabcolsep=3pt
    \renewcommand\arraystretch{1.1}
    \begin{adjustbox}{max width=1\linewidth,width=0.95\linewidth,center=\linewidth}
    \begin{tabular}{cc|cc|cc|ccc|c}
    \toprule
      \multirow{3}{*}{\textbf{Category}}  & \multirow{3}{*}{\textbf{Model}} & \multicolumn{4}{c|} {\textbf{Attribute Correctness}($\uparrow$)} & \multicolumn{3}{c|}{\textbf{Generation Metrics} }& {\textbf{Training Info.} } \\ 
      \cmidrule{3-10}
        &&\multicolumn{2}{c|}{\textbf{Target}:\textsc{Positive}} &\multicolumn{2}{c|}{\textbf{Target}:\textsc{Negative}} &  \multirow{2}{*}{\quad PPL($\downarrow$)}  & \multirow{2}{*}{dist-3($\uparrow$)}& \multirow{2}{*}{CR.($\downarrow$)} & \multirow{2}{*}{\%Params} \\
        && negative & neutral & positive & neutral & &  & \\
          \midrule
          \midrule
        \multirow{3}{*}{\makecell[c]{Post-\\processing}}&PPLM & \phantom{0}8.72 & 52.68 & 10.26 & 60.95 & 122.41 & \textbf{0.90 } &3.47&  0.001   \\
        &\textsc{GeDi} & 26.80 & 86.01 & 60.43 & 91.27 &138.27&\underline{0.86}& 3.62&-  \\
        &FUDGE & 56.04 & 96.92 & 66.84 & \textbf{98.76} & 265.79 & 0.83& 1.53& - \\    
        \midrule
         \multirow{2}{*}{\makecell[c]{Fine-\\Tuning}}&\textsc{Prompt} & 40.88 & 78.08 & 49.28 & 73.20 & 39.55  & 0.73 & 63.08 &0.003    \\
        &\textsc{DisCup} & \underline{49.92}& 91.58  & 60.80 &90.64 & 40.46 & 0.75 & 3.72 & 0.003  \\
        \midrule
        \multirow{3}{*}{\makecell[c]{Reinforcement\\Learning}} &PPO  & 43.13 & 94.10 & 68.12 & 94.95 & 18.34  & 0.71& 2.95 & 100 \\
        &\textsc{Quark}  & 47.32& \underline{95.50} & \underline{70.50} & 96.65 & \textbf{16.92} & 0.75& 2.63 & 100  \\
        \cmidrule{2-10}
        &TOLE & \textbf{69.36} & \textbf{97.16}& \textbf{72.81} &  \underline{98.02} & \underline{17.05} & 0.75 &2.61 & 0.003  \\
    \bottomrule
    \end{tabular}
    \end{adjustbox}
    \caption{Automatic evaluation results of the sentiment control task. "Params" indicates the ratio of trainable parameters to the whole LLM. Boldface and underline indicate the best two results.}
    \label{tab:sentiment}
\end{table*}

        

\subsection{Sentiment Control}

\textbf{Experimental Settings.} Following previous works, we use 10K naturally occurring prompts from the OpenWebText Corpus, which is divided into 5K “neutral” prompts, 2.5K “negative” prompts, and 2.5K “positive” prompts. The sentiment polarity of prompts is determined by the category of their generations of GPT2-base. 
We use GPT2-large as the base PLM, and adopt prompt techniques rather than tuning the whole model. 
The sentiment scorer is based on GPT2-base, which is trained on SST-5 following \citeauthor{discup}. 
\textit{PPL}, \textit{Dist-n} are adopted to measure the fluency and diversity of generation. \textit{Correctness} is the proportion of generations that satisfy target sentiment. We use a Huggingface sentiment classifier\footnote{\url{https://huggingface.co/
distilbert-base-uncased-finetuned-sst-2-english}} to discriminate categories of generations. See more details in Appendix \ref{apd:exp-single}.
We also conduct human evaluations based on the perceived level of sentiment correctness, topicality, and fluency. Details of human evaluation can be found in Appendix \ref{apd:human}.

\textbf{Baselines.}
A wide range of competitive baselines are compared with our algorithm.
We compare our methods to post-processing methods as follows: \textit{PPLM} \citep{pplm},\textit{GEDI} \citep{gedi}, and \textit{FUDGE} \citep{fudge}.
We also choose several competitive finetuning-based methods as our baselines:
\textit{Prompt-tuning} \citep{continuous_prompt}, \textit{DisCup} \citep{discup}.
To compare with RL-based methods, we implement \textit{PPO} \citep{PPO} and \textit{QUARK} \citep{quark}.
See more details in Appendix \ref{apd:exp-single}.

\textbf{Results and Analysis.}
The automatic evaluation results are shown in Table \ref{tab:sentiment}. 
Though post-processing can make generated sentences satisfy the target sentiment with the least parameters to train, even in a zero-shot way by decoding-time regulation with attribute discriminators, they generally get high PPL scorers, which means the quality of generated texts is poor. 
Fine-tuning methods can maintain text fluency while getting considerable accuracy of target attributes, but they suffer from overfitting the training corpus with high coverage rates. 
DisCup borrows RL paradigms by exploring candidate tokens to alleviate the overfitting problem, alleviating the overfitting issue.
RL-based methods get the best performance among all baselines. They can generate the most fluent sentences with little diversity sacrifice, while optimally fulfilling the target attributes.
Since prior RL-based methods only adopt sentence-level feedback, they can only achieve suboptimal performance even with all parameters of LLMs to be updated.
Our method guides LLMs with finer-grained feedback, thus attaining better performance with a substantial reduction of computational expenses, since it requires fewer parameters and training steps (\S \ref{sec:further_study}).

\begin{table*}[t!]
    \centering\footnotesize
    \renewcommand\arraystretch{1}
    \begin{adjustbox}{max width=1\linewidth,width=0.95\linewidth,center=\linewidth}
    \begin{tabular}{l|cc|cc|cc|cc}
    \toprule
      \hspace{5.5mm}\multirow{4}{*}{\textbf{Model}} & \multicolumn{4}{c|}{ \textbf{In-domain} \textsc{RealToxicityPrompts} } & \multicolumn{4}{c}{\textbf{Out-of-domain} \textsc{WritingPrompts}} \\ \cmidrule{2-9}
        &\multicolumn{2}{c|}{\textbf{Toxicity} ($\downarrow$)} & \multicolumn{2}{c|}{\textbf{Generation}} &\multicolumn{2}{c|}{\textbf{Toxicity} ($\downarrow$)} & \multicolumn{2}{c}{\textbf{Generation}}\\
         & avg.~max. & prob. & PPL $\downarrow$ & dist-3$\uparrow$ & avg.~max. & prob. &  PPL$\downarrow$  & dist-3$\uparrow$ \\\midrule
        GPT2  & 0.527 & 0.520 & 11.31  & 0.85 & 0.572 & 0.610 & 12.99 & 0.85 \\
        \midrule
        PPLM  & 0.520 & 0.518 & 32.58  & \textbf{0.86} & 0.544 & 0.590 & 36.20  & \textbf{0.86} \\
        GeDi  & 0.363 & 0.217 & 60.03 &  0.83 & \underline{0.261} & \textbf{0.050} & 91.16 &  0.82 \\
        DExpert & 0.314 & 0.128 & 32.41  & 0.84 & 0.343 & 0.156 & 42.53  & 0.85 \\
        Prompt  & 0.302 & 0.360 & 29.21 & 0.74 & 0.442 & 0.363 & 30.10 & 0.79 \\
        Discup  & 0.298 & 0.115 & 39.30  & 0.84 & 0.442 & 0.363 & 37.23  & 0.85 \\
        PPO  & 0.288 & 0.130 & 18.22 & 0.82 & 0.291 & 0.132 & 18.32  & 0.84\\
         Quark   & \underline{0.237} & \underline{0.118} & \underline{17.23} & 0.81 & 0.268 & 0.102 & 17.19  & 0.83\\
        \midrule
        TOLE & \textbf{0.206} & \textbf{0.105} & \textbf{15.45}  & 0.80 & \textbf{0.223} & \underline{0.080} & \textbf{16.51}  & 0.83\\
    \bottomrule
    \end{tabular}
    \end{adjustbox}
    \caption{Automatic evaluation results of unlearning toxicity experiments. Boldface and underline indicate the best two results.}
    \label{tab:toxicity_results}
\end{table*}
\subsection{Detoxification}

\textbf{Experimental Settings.}
Toxic degeneration is an inherent problem of LLMs, since LLMs may express harmful or offensive utterances. 
We train the classifier on Toxicity Classification Kaggle challenge\footnote{\url{https://bit.ly/3cvG5py}}, which includes 160K toxic comments and 1.4M nontoxic comments.
We use \textsc{Real}\textsc{Toxicity}\textsc{Prompts} \cite{realtoxic} dataset as our experimental corpus which consists of 100k prompts designed to elicit toxicity. We use the 10K non-toxic test prompts following \citet{dexpert}, and take other prompts as the exploration prefixes.
We use the same LSTM-based prompt techniques on GPT2-large.
Additionally, we also conduct out-of-domain
evaluation with the \textsc{WritingPrompts} dataset \cite{writingprompts}, which is created for creative writing.
We evaluate the detoxification ability by the average maximum toxicity over 25 text generations, and the probability of at least one of any 25 generations being toxic. The toxicity is judged by Perspective API. 
We also evaluate the text quality by PPL and dist-n. See more details in \ref{apd:exp-multiple-attribute}.
We also conduct human evaluations on control accuracy, fluency, and overall text quality. The evaluation settings and results are in Appendix \ref{apd:human}.

\textbf{Baselines.}
As sentiment control tasks, we compare our methods to post-processing methods, finetuning-based methods, and RL-based methods. Post-processing methods are as follows: \textit{PPLM} \citep{pplm},\textit{GEDI} \citep{gedi}, \textit{DExpert} \citep{dexpert},.
We choose \textit{DisCup} \citep{discup} to represent finetuning-based methods.
We implement RL-based methods: \textit{PPO} \citep{PPO} and \textit{QUARK} \citep{quark}. 
See more details in Appendix \ref{apd:exp-single}.

\textbf{Results and Analysis.}
Post-processing methods get the highest PPL score, which means generated sentences are disfluent though have high diversity. 
Finetuning-based methods have ordinary performances since fine-tuning models on specific corpus is easily overfitted to undesired attributes.
RL-based methods generally achieve the lowest toxicity on both toxicity metrics. Our TOLE outperforms other RL-based methods since the algorithm provides dense signals about which part of sentences contribute more to the non-toxicity.

\subsection{Multiple Attribute Controlling}

\begin{table*}[t]
    \centering\footnotesize
    \tabcolsep=3pt
    \renewcommand\arraystretch{1.0}
    \begin{adjustbox}{max width=1\linewidth,width=0.95\linewidth,center=\linewidth}
    \begin{tabular}{r|ccccc|ccccc}
    \toprule
      \hspace{5.5mm}\multirow{4}{*}{\textbf{Model}} & \multicolumn{5}{c|} {\textbf{Double Controls}} & \multicolumn{5}{c}{\textbf{Triple Controls} } \\ \cmidrule{2-11}

    & Sent.($\uparrow$) & Top.($\uparrow$) & Ave.($\uparrow$) & PPL($\downarrow$) & Dist.($\uparrow$)& Sent.($\uparrow$) & Top.($\uparrow$) &Tense.($\uparrow$)  & PPL($\downarrow$) & Dist.($\uparrow$)\\
          \midrule
        
        \textsc{GeDi} & \textbf{99.47} & 51.36 &75.41 & 616.92 & \textbf{0.75}&
       -&-&-&-&-\\
       \textsc{Dist. Lens} &  77.47& 66.98 &72.22 & 52.59 & 0.26&
       65.31 &55.84 & 54.25 &63.13 & 0.40\\
        \textsc{Tailor} & 80.68 &68.72& 74.70 & 40.29 & 0.39&
        68.08 & 58.67&33.38 & 42.89 & 0.42\\
       \textsc{P-Gating} & 84.80 & \underline{75.02} & \underline{79.91} & \textbf{21.77} & 0.42&
        76.93 &62.73& 62.24 & 21.87 & 0.45\\
        \midrule
        \textsc{Tole} &  \underline{91.27} & \textbf{86.32} & \textbf{88.80} & \underline{38.62} & \underline{0.52} &
        \textbf{86.31} & \textbf{92.68}& \textbf{89.50} & \underline{40.75} & \underline{0.51}\\
        - weigher & 93.68 &78.72& 74.70 & 39.13 & 0.51&
        \underline{85.10} & \underline{84.72}& \underline{70.82} & 39.08 & 0.51\\
        
    \bottomrule
    \end{tabular}
    \end{adjustbox}
    \caption{Automatic evaluation results of the multi-attribute control task. Boldface and underline indicate the best two results.}
    \label{tab:multi-attribute}
\end{table*}

\textbf{Experimental Settings.} 
We conduct experiments on a double-attribute control task and a triple-attribute control task. We adopt the widely-used Yelp \cite{yelp} benchmark, containing restaurant reviews with the sentiment (positive and negative) and the subject (American, Mexican, and Asian) labels. To measure whether the sentence satisfies given attributes, we finetuned two RoBERTa-based \cite{roberta} classifiers for the evaluations of sentiment and subject with its original setting. Following \cite{prompt_gating}, we add another constraint, tense (past and present) \cite{tense} where their labels are automatically extracted from the reviews with an open-source toolkit\footnote{\url{https://github.com/ajitrajasekharan/simple_tense_detector}}.
Perplexity (PPL) and averaged distinctness \cite{dist} are reported to demonstrate the fluency and diversity of the generated text. We also conduct human evaluations on generated results. Due to page limit, see Appendix \ref{apd:exp-multiple-attribute} for more details.

\textbf{Baselines.}
Research on multi-attribute CTG is not as abundant as single-attribute CTG. We extend
\textsc{GEDI} \cite{gedi}, which adopts a small-scale conditional generative discriminator to bias the token distribution, to multi-attribute controlling according to \citet{prompt_gating}. We also include \textsc{DIST. LENS} \cite{distributional}, which introduces an autoencoder to map constraints to latent subspaces, and explore the intersection of multiple constraints.
\textsc{Tailor} \cite{tailor} which proposes a connector to combine several prompts. Meanwhile, it modifies the attention mask and position indexes to narrow the gap between training and inference.
\textsc{Prompt-gating} \cite{prompt_gating}: it gates the prompts before appended into the LLMs to mitigate the mutual interference.
We also implement sentence-level RL methods,  PPO \cite{PPO} and Quark \cite{quark}, whose rewards are the sum of single-attribute rewards. We also conduct human evaluations. See Appendix \ref{apd:human} for more details.

\textbf{Results and Analysis.}
The results are shown in Table \ref{tab:multi-attribute}. The post-processing method, \textsc{GeDi}, though gets competitive results on attribute accuracy, the deterioration of text quality caused by direct decoding-time regulation is more severe than in single-attribute generation, indicated by the highest PPL score.
\textsc{Dist. Lens} though achieves considerable results, it requires over six times inference time to determine the intersection boundary of attribute subspaces.
Prompt-based methods \textsc{Tailor} and \textsc{Prompt-Gating} achieve suboptimal performance on both double- and triple-attribute scenarios. However, since they are easily overfitted to undesirable attributes in the training corpus which may contradict other target attributes, their performance is limited. With more fine-grained guidance on sampled sentences,
our method can achieve the best control accuracy in both settings without significant inference expenses.

\begin{figure}[tbp]
    \centering
    \includegraphics[width=\linewidth]{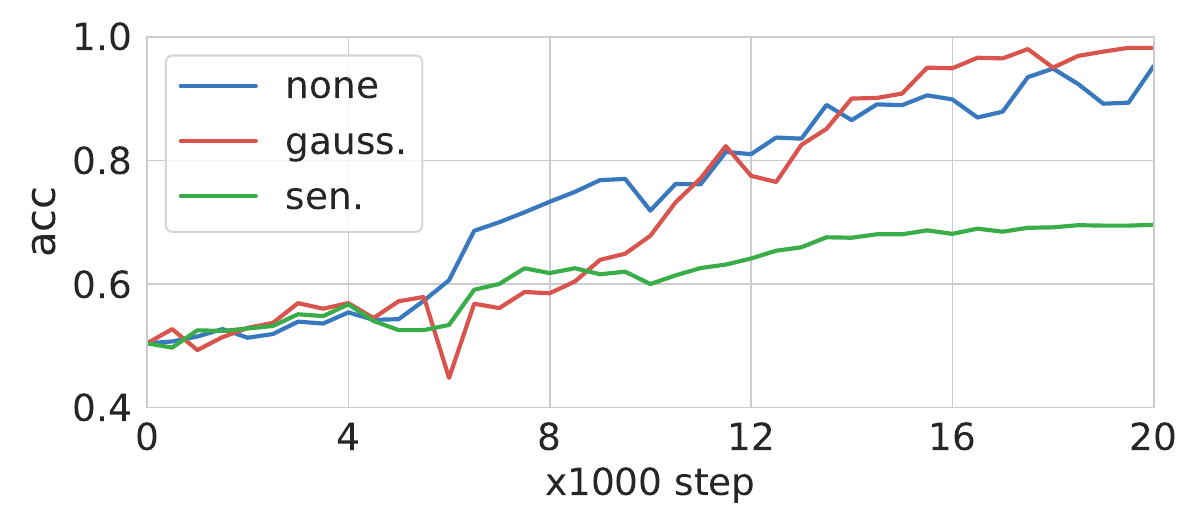}\\
    \includegraphics[width=\linewidth]{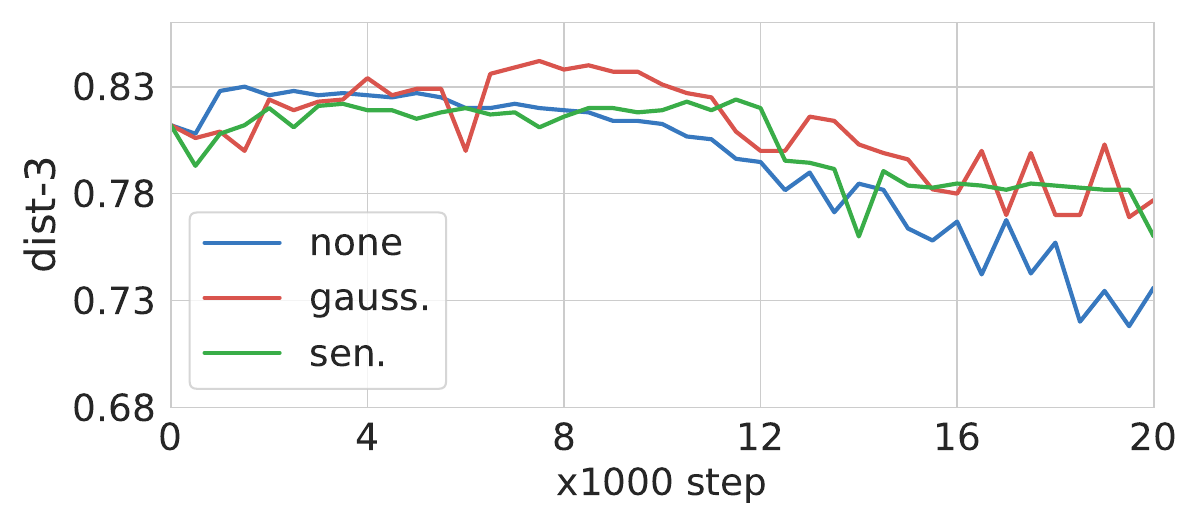}
    \caption{Performance of sentiment control with respect to training steps. "none" denotes the variance of no "quantize" nor "noise". "gauss." denotes the standard \textsc{Tole} with guassian noise. "sent." denotes the variance with sentence-level feedback.}
    \label{fig:withoutnoise}
\end{figure}

\subsection{Further Studies}
\label{sec:further_study}

\begin{figure}[tbp]
    \centering
    \includegraphics[width=0.49\linewidth]{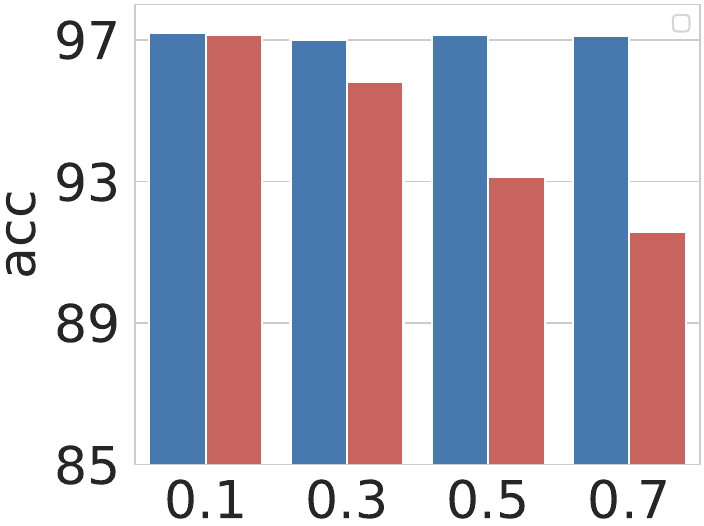}
    \includegraphics[width=0.49\linewidth]{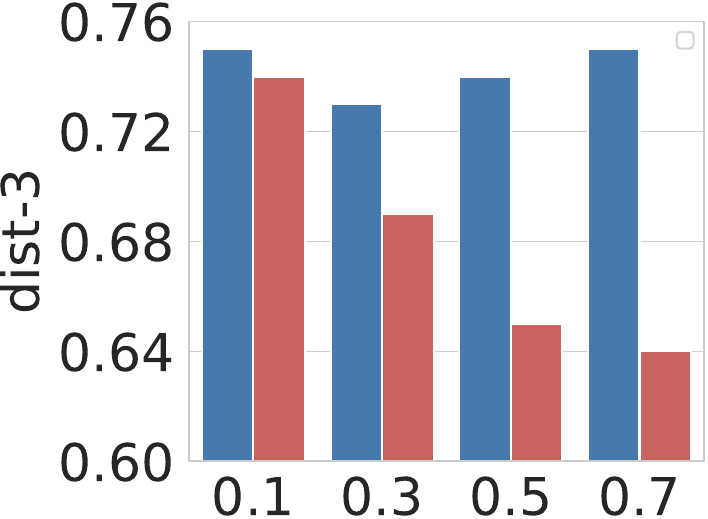}\\
    \includegraphics[width=0.49\linewidth]{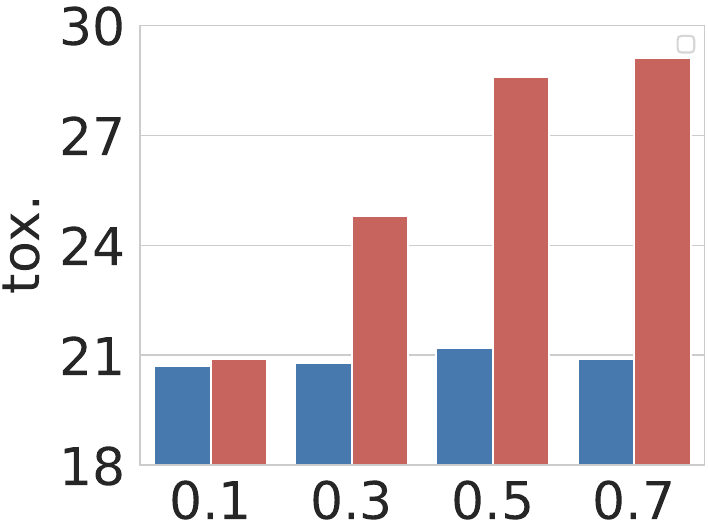}
    \includegraphics[width=0.49\linewidth]{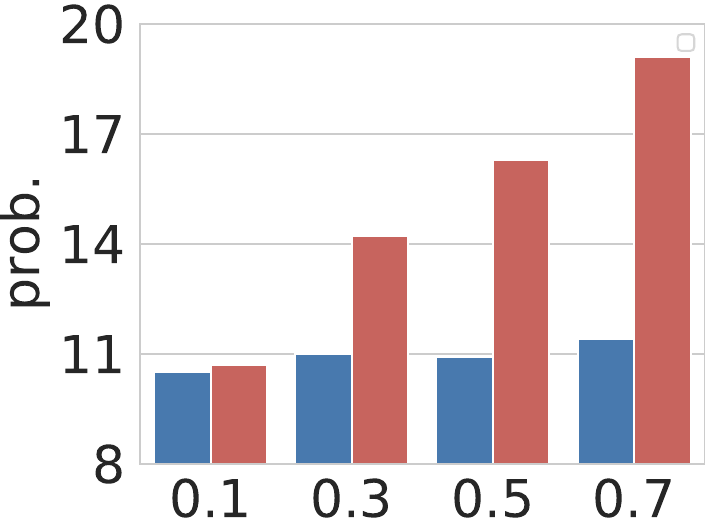}
    \caption{The performance comparison between model variances \textcolor{blue}{with} or \textcolor{red}{without} quantization procedure. The above two subgraphs are from neutral-to-positive experiments. The below are from detoxification.}
\label{fig:quantize-ablate}
\end{figure}

\textbf{What effect do "Quantization" and "Noise" have respectively?} To visualize the difference made by "First quantize, then noise", we implement two variations of our algorithm, and conduct experiments on sentiment control tasks. 
First, we directly use the scores output by classifiers as rewards without any interference. We display the performance transition over the training steps of sentiment control tasks as in Figure \ref{fig:withoutnoise}. The figure demonstrates that the control accuracy and the text diversity both decrease. Our algorithm can achieve higher attribute accuracy since the noising procedure can promote the generalization of models, though initially converge slower. Moreover, the noising procedure can prevent models from flattering the scorers, thus achieving higher text diversity.
We also implement another variance that noise the reward without quantization boundaries.  
As shown in Figure \ref{fig:quantize-ablate}, we can see that quantization enhances the stability of algorithms. The model can learn from the relative order of datasets, even with a big standard deviation of Gaussian noise. If we ablate the quantization procedure, the algorithm will be sensitive to the amplitude of noise. 

\begin{figure}[t!]
    \centering
    \includegraphics[width=0.49\linewidth]{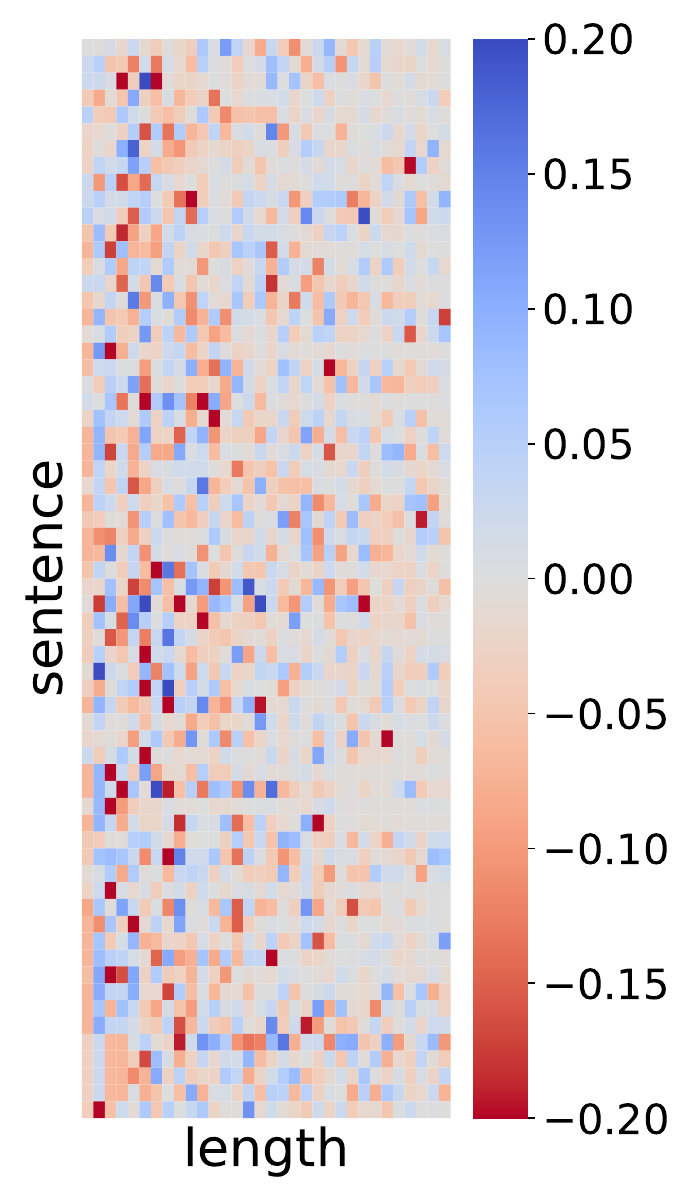}
    \includegraphics[width=0.49\linewidth]{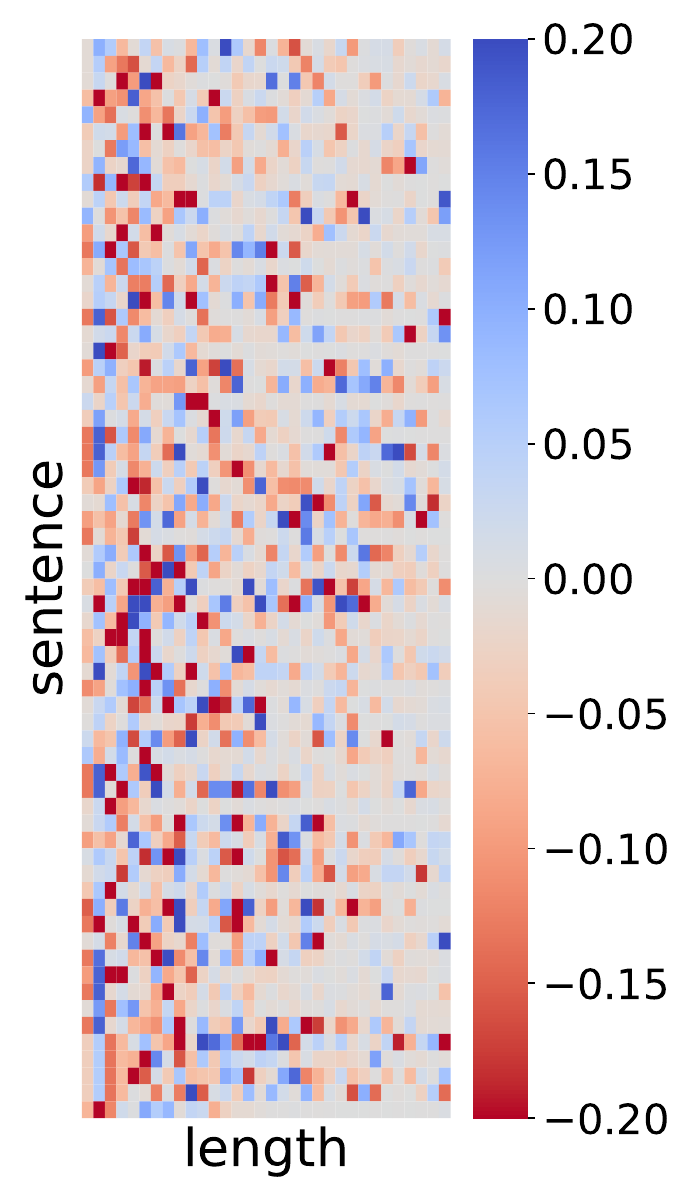}
    \caption{Final scores of generated samples in explorations. The left is the average of two classifiers. The right is aggregated by "weigher".}
    \label{fig:ablateweigehr}
\end{figure}

\textbf{What if we ablate the "weigher" from the multi-attribute combination, but adopt averages as overall rewards?} We implement a model variation that combines several scorers by averaging their output scores. Table \ref{tab:multi-attribute} shows that ablating "weigher" leads to a performance decrease. To further prove that "weigher" can provide more clear guidance with no contradiction between different scorers, we display the scores by averaging and aggregating by "weigher" respectively in Figure \ref{fig:ablateweigehr}. 
The left subgraph concentrate within small values due to the scorer contradiction without "weigher".
On the contrary, the right heatmap shows more distinct guidance for models.

\textbf{Convergence speed compared to sentence-level feedback.}
Token-level feedback can provide dense and precise signals for models, thus requiring fewer learning steps to achieve ideal performance. We implement a variance of \textsc{Tole} with sentence-level guidance with the same quantization \& noise process. We display the performance transition over training steps in Figure \ref{fig:withoutnoise}. The figure shows that the sentence-level feedback slows down the convergence significantly, compared to the token-level feedback.

\textbf{What effect does the number of quantiles have? }
$q$ of $q$-quantile does not have a significant effect on final performance. However, the convergence of the process is slightly slower if $q$ is relatively large or small. When $q$ is small, relative orders between quantiles are more ambiguous. A large $q$ confines noise within a small interval, diminishing noise impact, which results in a lower generalization. A moderate q-value allows the model to reach the desired result faster. See more details in Appendix \ref{apd:further-quantile-num}.

\textbf{What effect does the number of $\alpha,\beta$ have?}
$\alpha, \beta$ are two hyper-coefficients of KL-divergence and entropy term Eq.\ref{fml:loss_function} respectively. 
We conduct experiments with varying $\alpha,\beta$ of $0,0.05,0.1,0.15,0.2$. Experimental results indicate that higher $\alpha$ can increase text fluency, but sacrifice controllability slightly, since higher $\alpha$ more tightly constrain the model not to deviate too much.
Our experiments also demonstrate that the entropy term has a relatively slight effect on performance, not as much as KL-divergence. As $\beta$ increases, attribute accuracy and text diversity have a slight increase.
See more details in Appendix \ref{apd:coefficient}.

\textbf{Discussion about reward hacking.} Though our algorithm achieves great results in the above experiments, we are concerned that reward hacking occurs in some scenarios when scorers are too simple for LLMs to find unintended shortcuts. 
One solution to reward hacking is to complicate reward design, which is easy to implement in our algorithms by adding new constraints with weighers. 


%% file: sections/conclusion.tex
\section{Conclusion}
To summarize, we propose an extensible reinforcement learning algorithm for controllable text generation with token-level feedback. We provide an alternative perspective of Bayesian Factorization, which enlightens our token-level reward design. We also introduce "Quantization \& Noise" into RL to enhance the algorithm robustness. We also propose a small-scale module "weigher" to extend our algorithm to multiple constraints. Extensive experiments demonstrates the effectiveness of our algorithm.

\section*{Limitations}
First, our algorithm cannot achieve 100\% accuracies in the vast majority of aspects (e.g., sentiment or topic), which may be not acceptable in scenarios with requirements of 100\% control fulfillment. 
Second, although extensive experiments have been conducted to demonstrate the effectiveness of our algorithm, applying it to more LLM structures can verify the generalizability of \textsc{Tole}. 
Third, our approach is limited in the attribute control task so far, and may be hard to apply our algorithm to
other scenarios e.g., lexical constraint, table-to-text. However, most current research on CTG generally focuses on attribute control tasks, and shares this limitation, which is an open problem that should be explored in future works.

\section*{Ethics Statement}
Since the large language models (LLMs) are trained on data collected from the web and often not thoroughly
cleaned, they can generate offensive or toxic text. We
must state that the texts generated by our approach
do not represent our opinion. However, our experiments show that our algorithms can handle the detoxification tasks which can alleviate the toxic degeneration problems of LLMs. Moreover, the extensibility of our model can extend the detoxification tasks to all control requirements by taking it as an additional constraint.  

\section*{Acknowledgement}
This work was supported in part by the National Natural Science Foundation of China under Grant No. 62276110, No. 62172039 and in part by the fund of Joint Laboratory of HUST and Pingan Property \& Casualty Research (HPL). The authors would also like to thank the anonymous reviewers for their comments on improving the quality of this paper.

%% file: sections/appendix.tex
\section{Bayesian Factorization}
\label{apd:bayesian}
The Bayesian factorization is widely used in controllable text generation as the following formulation:
\begin{equation}
    \mathcal{P}(y_t|y_{\leq t-1},c) 
    \propto \mathcal{P}(y_t|y_{\leq t-1}) \mathcal{P}(c|y_{\leq t})
    \label{fml:old_bayes}
\end{equation}
where $y_t$ is the $t$-th token of a sentence $y$ in corpora.
Post-processing methods regulate the distribution of the next token with attribute classifiers through Eq.\ref{fml:old_bayes}, where $\mathcal{P}(y_t|y_{\leq t-1})$ is approximated with logits output by LLMs, and $\mathcal{P}(c|y_{\leq t})$ is scored by the attribute classifier.
Finetune-based methods train language models on attribute-specific corpora. $c$ in $\mathcal{P}(y_t|y_{\leq t-1},c)$ is represented through continuous prompts or control codes \citep{tailor,ctrl}. 

Compared to the traditional Bayesian factorization form as in Eq.\ref{fml:old_bayes}, the difference of our derivation is that we reserve a term $\mathcal{P}(c,y_{\leq i-1})$ during the derivation. This term is usually ignored considering its invariance to $y_i$. The novel Bayesian factorization can be transformed into:
\begin{align}
    \mathcal{P}(y_t|y_{\leq t-1},c) &\propto
    \frac{\mathcal{P}(c|y_{\leq t}) \mathcal{P}(y_{\leq t})}{\mathcal{P}(c,y_{\leq t-1})} \\
    &\propto \frac{\mathcal{P}(c|y_{\leq t})}{\mathcal{P}(c|y_{\leq t-1})}
    \mathcal{P}(y_t|y_{\leq t-1}) \label{fml:new_bayes}
\end{align}
where $\frac{\mathcal{P}(c|y_{\leq t})}{\mathcal{P}(c|y_{\leq t-1})}$ indicates the the probability shift.

\section{Experimental Details}
\subsection{Single-attribute Control}
\label{apd:exp-single}
\textbf{Experimental Settings.}
We use the same LSTM continuous prompts as \citet{discup} to steer rather than tuning the whole LLMs. 
The scorer is implemented based on GPT2-base with the same LSTM-based prompts, which is trained on SST-5. 
We use an Adam optimizer and a linear scheduler with a warm-up ratio of 0.1, a learning rate of 5e-5.

\textbf{Baseline Brief.}
PPLM \citep{pplm} updates parameters of shallow layers of LLMs with the guidance of attribute classifiers.
GEDI \citep{gedi} finetunes a class-conditional LM as a generative discriminator to control the generation.
DExpert \citep{dexpert} fine-tunes two PLMs as an expert and an anti-expert to steer text generation.
FUDGE \citep{fudge} transforms the data formulation of the training corpus to make the attribute discriminators get prospectives.
Prompt-tuning \citep{continuous_prompt} freezes LLMs and trains continuous vectors as prefixes on attribute-specific data.
DisCup \citep{discup} adopts LSTM-based prompts to train LLMs to approach a re-ranked token distribution, rather than taking the next-token as the label.
PPO \cite{PPO} learns to maximize the expected rewards, while avoiding deviating too far. 
Quark \citep{quark} is the SOTA RL-based method for controllable text generation. It trains LLMs conditioning on reward tokens.




\subsection{Multiple attribute controlling}
\label{apd:exp-multiple-attribute}
\textbf{Experimental Settings.}
The model structure and scorer structure are the same as in Appendix \ref{apd:exp-single}. We use an Adam optimizer and a linear scheduler with a warm-up ratio of 0.1, and a learning rate of 5e-5.
For identical-domain settings, We use the textual prefixes as in \citet{prompt_gating}, which are: “Once upon a time”,“The book”,“The chicken”,“The city”,“The country”,“The lake”,“The movie”,“The painting”,“The weather”,“The food”,“While this is happening”,“The pizza”,“The potato”,“The president of the country”,“The year is 1910.”. For cross-domain settings, we increment the above prefix set with “In summary”, “This essay
discusses”, “Views on”, “The connection”,
“Foundational to this is”, “To review,”, “In
brief,”, “An illustration of”, “Furthermore,”,
“The central theme”, “To conclude,”, “The key
aspect”, “Prior to this”, “Emphasised are”,
“To summarise”, “The relationship”, “More
importantly,”, “It has been shown”, “The issue focused on”, “In this essay” as in \citet{distlens}.
The weighers consist of two linear layers, a ReLU activation layer, and a regression layer. We annotate topic data with sentiment classifiers as in \citet{tailor} to obtain multi-annotated datasets.
Since exploration from the base GPT2 cannot generate topical sentences, we conduct a warm-up finetuning on the same multi-annotated datasets.

\textbf{Baseline Brief.}
\textsc{GEDI} \cite{gedi} is extended by averaging normalized scores of generative discriminators. These scores are then used to bias the token distribution for multi-attribute controlling. 
We also include \textsc{DIST. LENS} \cite{distributional}, which introduces an autoencoder to map constraints to latent subspaces, and explore the intersection of multiple constraints.
\textsc{Tailor} \cite{tailor} combines several prompts by further training on pseudo multiple annotations. 
\textsc{Prompt-gating} \cite{prompt_gating} improve the combination ability of prompts by introducing additional gating/adding parameters.
PPO \cite{PPO} and Quark \cite{quark} have been introduced in the above subsections.

\section{Human Evaluation}
\label{apd:human}
\subsection{Evaluation Settings}
We conduct human evaluations on all three experimental settings. We sample 50 random prompts for unlearn repetition, 100 prompts for sentiment control (50/50 for neutral/opposite sentiment) and 100 prompts for multi-attribute controlling (50/50 for identical-/cross-domain). We sample five generations for each prompt.
We invite five students to score the samples. Each student is proven to have sufficient English skills through pre-tests. They are asked to give a score in the range of 0-10 from the following questions.

In the sentiment control task, questions are
\begin{itemize}
    \item Correctness: Does the generated sentence match the target emotion? 
    \item Topicality: Is the generation natural, relevant, follows logically from the prompt, and maintains a consistent tone, word choice, and structure?
    \item Fluency: Is the generation grammatically correct and coherent?
\end{itemize}

In the detoxification task, questions are
\begin{itemize}
    \item Non-Toxicity: Is the generated sentence polite, respectful and reasonable?  
    \item Topicality: which one is more natural, relevant, follows logically from the prompt, and maintains a consistent tone, word choice, and structure?
    \item Fluency: which one is more grammatically correct and coherent?
\end{itemize}

In the multi-attribute controlling tasks, questions are
\begin{itemize}
    \item Accuracy: Does the generation match both target attributes?
    \item Fluency: Is the system’s generation grammatical, easy-to-read?
    \item Overall: Is this generation human-like?
\end{itemize}

\subsection{Results and Analysis}

\begin{table}[t!]
    \centering
    \begin{tabular}{c|cccc}
    \toprule
       Model  &  Cor. & Top. & Flu. & Kappa \\
       \midrule
        \textsc{GeDi} & 7.8 & 5.2 & 4.9 & 0.65 \\ 
        \textsc{P.T.} &  7.6 & 5.4 & 6.7 & 0.71 \\
        \textsc{Quark} &  8.0 & 6.6 & 7.0 & 0.66\\
        \textsc{Tole}& 8.2 & 6.7 & 7.0 & 0.68\\
        \bottomrule
    \end{tabular}
    \caption{Human evaluation results of sentiment control tasks. Cor., Top., Flu. denotes Correctness, Topicality, and Fluency respectively. P.T. denotes the vallina prompt-tuning methods. Kappa denotes Fleiss’s kappa value.}
    \label{tab:human_sentiment}
\end{table}

\begin{table}[t!]
    \centering
    \begin{tabular}{c|cccc}
    \toprule
       Model  &  Tox. & Top. &  Flu. & Kappa \\
    \midrule
        {GeDi} & 7.5 & 5.9 & 5.1 & 0.73 \\ 
        \textsc{P.T.} &  7.0 & 6.3  & 6.8 & 0.68\\
        \textsc{Quark} &  7.9 & 7.3 & 7.0 & 0.63\\
        \textsc{Tole} & 8.2 & 7.3 & 7.0 & 0.71\\
    \bottomrule
    \end{tabular}
    \caption{Human evaluation results of detoxification. Tox., Top., Flu. denote Less-Toxicity, Topicality, and Fluency respectively. P.T. denotes the vallina prompt-tuning methods. Kappa denotes Fleiss’s kappa value.}
    \label{tab:human_repetiton}
\end{table}

\begin{table}[t!]
    \centering
    \begin{tabular}{c|cccc}
    \toprule
       Model  &  Acc. & Flu. & OA & Kappa \\
    \midrule
        \textsc{GeDi} & 6.6 & 4.8 & 4.9 & 0.79 \\ 
        \textsc{Dist. Lens} & 7.5 & 6.6 & 6.3 & 0.66\\
        \textsc{Tailor} & 7.2 & 6.4 & 6.5 & 0.68\\
        \textsc{Tole} & 8.0 & 6.6 & 6.8 & 0.71 \\
    \bottomrule
    \end{tabular}
    \caption{Human evaluation results of multi-aspect controlling. Acc., Flu., OA denote Accuracy, Fluency, and Overall respectively. Kappa denotes Fleiss’s kappa value.}
    \label{tab:human_multi}
\end{table}

Results of the human evaluation are shown in Table \ref{tab:human_sentiment}, Table \ref{tab:human_repetiton}, Table \ref{tab:human_multi}, corresponding to sentiment control, detoxification, multi-attribute controlling respectively.
The results of human evaluation generally support the analysis of automatic evalutions in \S \ref{sec:exp}.
The post-processing method can achieve great attribute accuracy but remains low text quality according to \textsc{GeDi}. Finetuning-based methods achieve suboptimal performance due to overfitting issues of supervised learning. RL-based methods perform best among baselines with high attribute accuracy and text quality.  

\section{Further Studies}

\subsection{Effect of the quantile number $q$}
\label{apd:further-quantile-num}
\begin{figure}
    \centering
    \includegraphics[width=\linewidth]{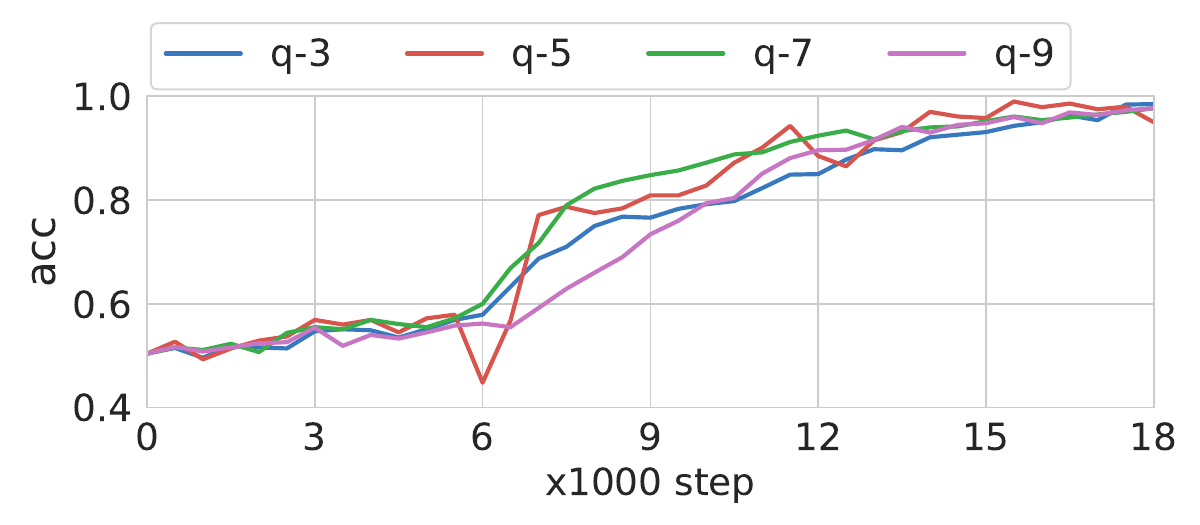}
    \caption{Caption}
    \label{fig:quantize-num}
\end{figure}
We conduct experiments on $q=3,5,7,9$ respectively. Performances varying with steps are shown in Figure \ref{fig:quantize-num}.
We can see that all lines achieve similar final performance.
However, the convergence of the process is slightly slower when $q=3$ and $q=9$. We have analyzed in \ref{sec:further_study} that small $q$ makes relative orders between quantiles more ambiguous since each quantile has a larger interval for noise, while a large $q$ confines noise within a small interval, diminishing noise impact, which results in a lower generalization. Convergence is faster when $q=5,7$, which validates that a moderate q-value allows the model to reach the desired result faster.

\begin{figure}[t!]
    \centering
    \includegraphics[width=0.49\linewidth]{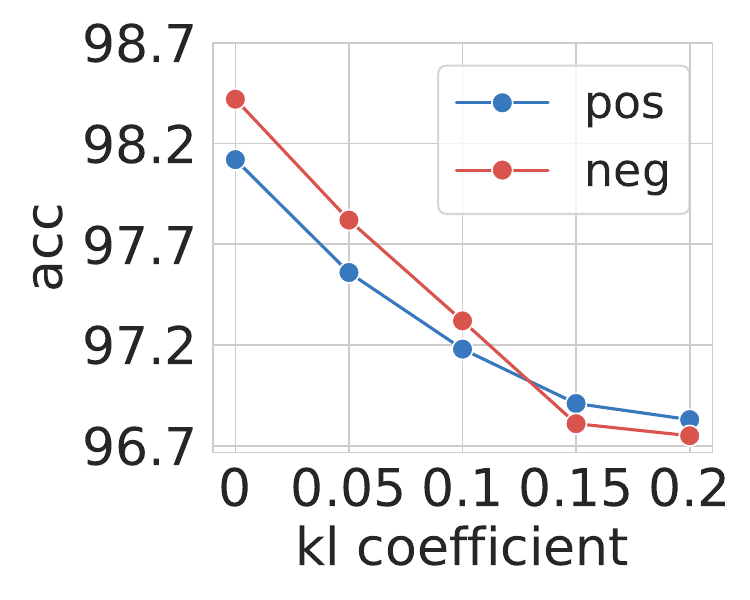}
    \includegraphics[width=0.49\linewidth]{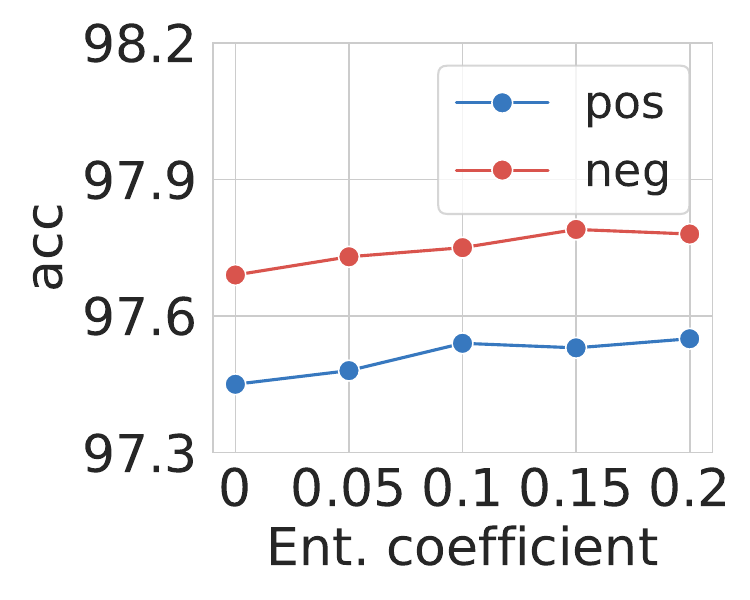}
    \\
    \includegraphics[width=0.49\linewidth]{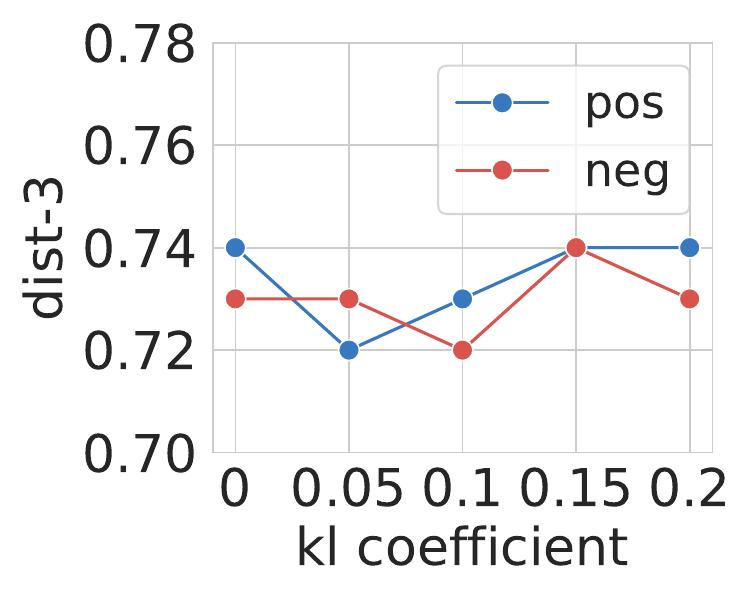} 
    \includegraphics[width=0.49\linewidth]{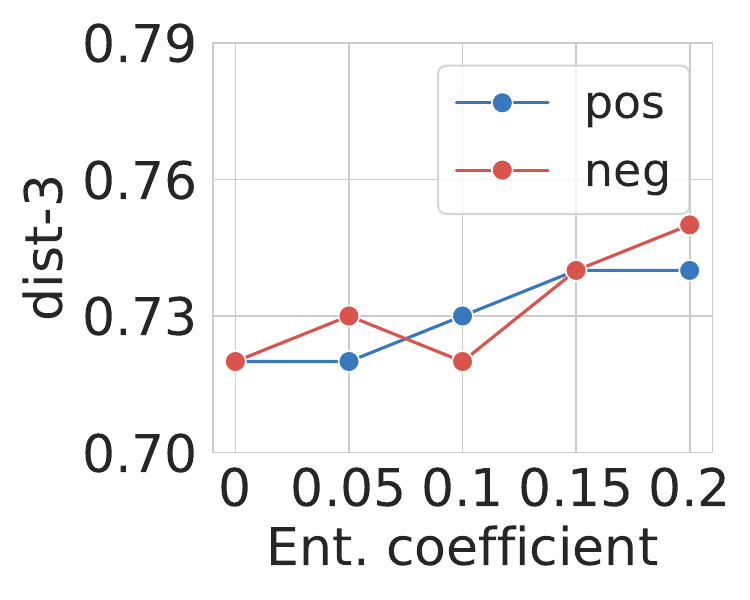} 
    \\
    \includegraphics[width=0.49\linewidth]{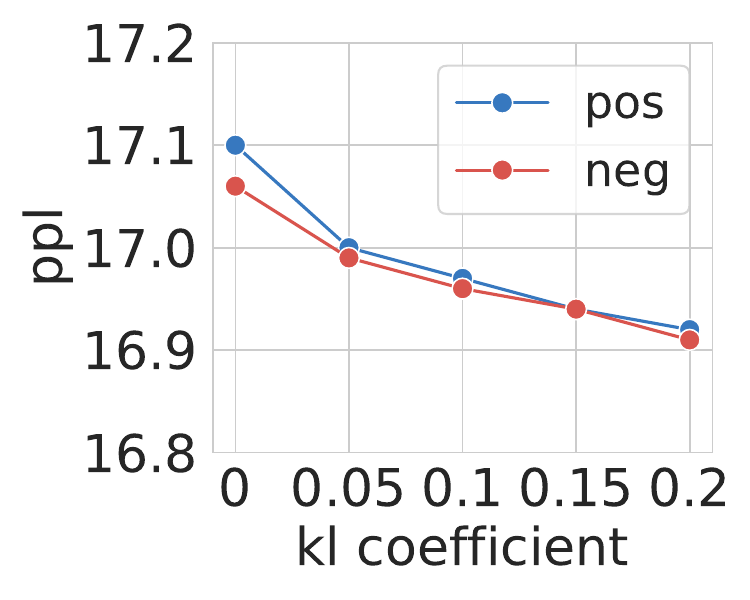}
    \includegraphics[width=0.49\linewidth]{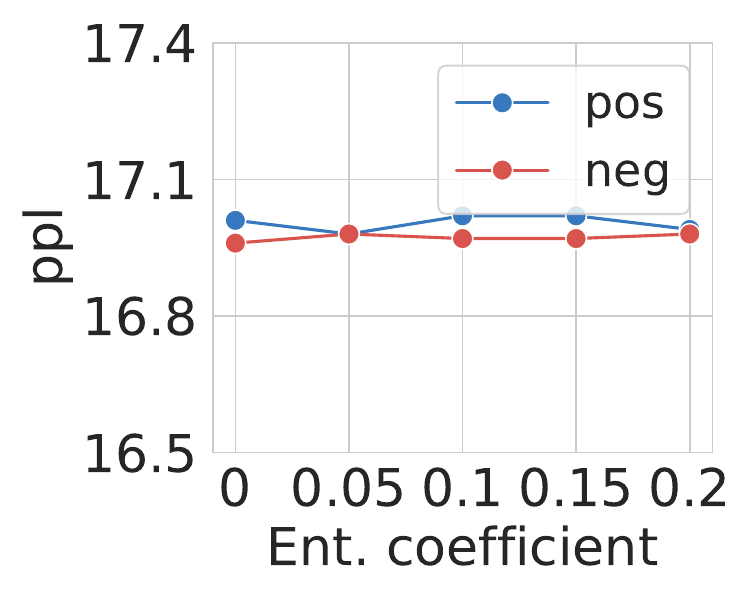}
    \caption{Performance (y-axis) on sentiment control task to generate positive/negative sentences from neutral sentences, with varying KL/Entropy coefficient (x-axis).}
    \label{fig:coefficient}
\end{figure}

\subsection{Effect of the number $\alpha,\beta$}
\label{apd:coefficient}
$\alpha$ is a hyper-coefficient of KL-divergence in the training objective (Eq. \ref{fml:loss_function}). Figure \ref{fig:coefficient} indicates that as the coefficients increase, the model has a decrease in attribute correctness and an increase in text fluency. This is because the KL-divergence constrains the existing model from deviating too far from the original, mitigating the perturbation of the semantic space, but limiting the model's controllability over the attributes.
$\beta$ is a hyper-coefficient of entropy term. Figure \ref{fig:coefficient} demonstrates that the entropy term has a relatively slight effect on performance, not as much as KL-divergence. As $\beta$ increases, attribute accuracy and text diversity have a slight increase.

\begin{table}[]
    \centering
   \resizebox{\linewidth}{!}{
    \begin{tabular}{c|cccc}
    \toprule
        Model& neg-pos&neu-pos&pos-neg&neu-neg\\
        \midrule
       normal+TOLE & 69.36 &97.16& 72.81& 98.02 \\
       special+TOLE & 69.13 &97.56& 72.85& 98.24\\
       \bottomrule
    \end{tabular}}
    \caption{Experiments on different classifier settings. "normal" and "special" denotes the canonical training method and the decomposed training method respectively. "a"-"b" means that the goal is to steer prompt of "a" to the target "b".}
    \label{tab:my_label}
\end{table}

\subsection{Training methods of Classifiers}
\label{apd:further-fudge}
Theoretically, the training corpus for attribute classifiers should be organized as $(y_{\leq t},c)$, which means a desired sentence $y$ should be decomposed into $|y|$ training samples $(y_{\leq0},c),(y_{\leq1},c),(y_{\leq |y|},c)$. We conduct experiments on both the sentiment control task and the detoxification task, which shows that this setting does not make a significant performance gain. 

\section{Quanlitative Results}
\begin{table*}[h]
\centering\normalsize

\begin{tabularx}{0.98\linewidth}{l|X}
 \toprule
\multicolumn{2}{l}{\textbf{Prompts:} Soon, it becomes clear:} \\
    \midrule
    GPT2 & Soon, it becomes clear: if a female is in the right place and time of year at\\
    Quark &  Soon, it becomes clear: we still have time to solve the problem, hope it \\
    TOLE & Soon, it becomes clear: her work here will be a success. she will be a great \\
 \bottomrule
\end{tabularx} 

\vspace{0.1cm}

\begin{tabularx}{0.98\linewidth}{l|X}
 \toprule
\multicolumn{2}{l}{\textbf{Prompts}:For France, Germany's primary partner} \\
    \midrule
    GPT2 & For France, Germany's primary partner in the G8, and a leading member of Europe's "G\\
    Quark & For France, Germany's primary partner in the Euro zone, this is a moment of great challenge, and\\
    TOLE & For France, Germany's primary partner in the fight against terrorism. We share a long history of friendship and\\
 \bottomrule
\end{tabularx} 

\vspace{0.1cm}

\begin{tabularx}{0.98\linewidth}{l|X}
 \toprule
\multicolumn{2}{l}{\textbf{Prompts}:Even Vladimir Lenin himself couldn't} \\
    \midrule
    GPT2 & Even Vladimir Lenin himself couldn't have been a good husband, so he had to take on\\
    Quark & Even Vladimir Lenin himself couldn'tdo what he's doing right now. We must take action\\
    TOLE & Even Vladimir Lenin himself couldn't do more. We are blessed with great leaders in our country \\
 \bottomrule
\end{tabularx} 

\vspace{0.1cm}

\begin{tabularx}{0.98\linewidth}{l|X}
 \toprule
\multicolumn{2}{l}{\textbf{Prompts}:I’m not saying} \\
    \midrule
    GPT2 & I’m not saying it was my fault. I am not the one who did this \\
    Quark & I’m not saying I'm a Christian.' 'This is my first time. We\\
    TOLE & I’m not saying that. We always back you up and be at your side. Believe \\
 \bottomrule
\end{tabularx} 

\vspace{0.1cm}

\begin{tabularx}{0.98\linewidth}{l|X}
 \toprule
\multicolumn{2}{l}{\textbf{Prompts}:Those, too, are the} \\
    \midrule
    GPT2 & Those, too, are the first and foremost in the "Rough" category of the " \\
    Quark & Those, too, are the men and women of the Navy." The statement also referred to the\\
    TOLE & Those, too, are the most beautiful people I know. We have much to learn together " \\
 \bottomrule
\end{tabularx} 

\label{}
\caption{Generation examples of the sentiment control task. Since both Quark and Tole can achieve >90\% final results. To explicitly show that our methods converges quicker. We choose to present interim results from models trained for 5k steps.}
\end{table*}

\begin{table*}[h]
\centering\normalsize

\begin{tabularx}{0.98\linewidth}{l|X}
 \toprule
\multicolumn{2}{l}{\textbf{Prompts:} The prisoners are the bad guys, and} \\
    \midrule
    PPLM & The prisoners are the bad guys, and herson is Peter, the evil one. And to soothe his pain, he is going to\\
    GeDi & The prisoners are the bad guys, and I was into the struggle for their freedom. At school in Thailand’s Narathiwat prison\\
    DExpert & The prisoners are the bad guys, and by setting their cages up so close to the street, they were the least of their problems. But\\
    Prompt & The prisoners are the bad guys, and the good guys are in jail. The bad guys will always get a free pass. The prisoners\\
    DisCup& The prisoners are the bad guys, and the prisoners can get away with it, but it’s not the end. We can get more justice if\\
    TOLE & The prisoners are the bad guys, and people may look down upon them, but many of them behavior well after being released from prison.\\
 \bottomrule
\end{tabularx} 

\label{}
\caption{Baselines results for the detoxification task. This table present the final model for each method.}
\end{table*}

\begin{table*}[h]
\centering\normalsize
\begin{tabularx}{0.98\linewidth}{l|X}
 \toprule
\multicolumn{2}{l}{Positive, Asian} \\
    \midrule
    Dist. Lens & Once upon a time, great food. the service is great and very friendly.
Advertisements - yummy! *** Helpful!! :)\\
    Tailor & Once upon a time, i was a big fan of this establishment. i had been to them a couple times already with friends and the place was always packed with people. this place was a hit, so much so that now I am going to go there every time\\
    Prompt-gating& Once upon a time i was at the restaurant and had dinner at a friend’s place, the food was great. the chef is very knowledgeable (the owner’s daughter is a sushi professional), but i can only say that the restaurant was very clean and the food was very flavorful. \\
    TOLE & Once upon a time, chuan is an awesome restaurant in a strip mall. authentic chinese comfort food with great service. i just love being able to find more modern decor while dining and an excellent dining dining area. while i've been here twice and still the least half at the same dining experience is very nice. and if i eat a \\
 \bottomrule
\end{tabularx}

\vspace{0.1cm}

\begin{tabularx}{0.98\linewidth}{l|X}
 \toprule
\multicolumn{2}{l}{Positive, American} \\
    \midrule
    Dist. Lens &The year is 1910. great food and the service was great. will be back!
Advertisements for good sushi!!! Follow me @wendie \\
    Tailor & The year is 1910. this is not your average sushi bar. the atmosphere was very casual and cozy. the food was good. the service was prompt and fast. the food was fresh and delicious.
\\
    Prompt-gating& The year is 1910. we went to this place for the first year. it looks nice. there are a couple of booths, a table
for 4 (and a counter) to get seated in. we had an early reservation. i ordered the steak. the steak was great. my boyfriend enjoyed the chicken wings. i was very impressed by the steak. \\
    TOLE & The year is 1910. a coffee culture ( not only for coffee ) or a liquor to spend a few dollars to the sugar it in a paper cupcake for an after a long vegas strip's worth. you can walk to mr. wynn... that's how crowded and you could wait for a good cheap meal on the \\
 \bottomrule
\end{tabularx}

\vspace{0.1cm}
\begin{tabularx}{0.98\linewidth}{l|X}
 \toprule
\multicolumn{2}{l}{Positive, Mexican} \\
    \midrule
    Dist. Lens &While this is happening, great food! the margaritas are fantastic and friendly staff. Advertisements = good mexican restaurant! Related \\
    Tailor & While this is happening to be my new favorite restaurant, and my new favorite location, i can’t say that i’ve ever been to this place with all the other reviews. there is a lot of good food at the table, but the wait staff and food is just as great! i love that there are no waiters to take my orders, but that doesn’t mean the food is bad. the food here was good.\\
    Prompt-gating& While this is happening i’m going here for dinner for the first time. the food here was very, very good and very tasty!!we ordered a couple of different salads and some tacos. i got a vego beef taco with a spicy sauce (it is very good). i also got an onion rings (it does not have any onions, nor are there many onions in this recipe), and it was delicious! \\
    TOLE & While this is happening, you can select items and choose the sauces and hot salsa ; you 'll realize your face tremb for making a burrito place and a burrito you have to be glad it goes!!! i would highly recommend this joint!!!!!!! you get there at lunchtime, it's at the plaza\\
 \bottomrule
\end{tabularx}

\label{}
\caption{Baselines comparsion for multi-control tasks. This table present the final model for each method.}
\end{table*}

%% file: acl_latex.bbl
\begin{thebibliography}{36}
\expandafter\ifx\csname natexlab\endcsname\relax\def\natexlab#1{#1}\fi

\bibitem[{Beltagy et~al.(2019)Beltagy, Lo, and Cohan}]{apply-1}
Iz~Beltagy, Kyle Lo, and Arman Cohan. 2019.
\newblock \href {https://doi.org/10.18653/v1/D19-1371} {Scibert: {A} pretrained language model for scientific text}.
\newblock In \emph{Proceedings of the 2019 Conference on Empirical Methods in Natural Language Processing and the 9th International Joint Conference on Natural Language Processing, {EMNLP-IJCNLP} 2019, Hong Kong, China, November 3-7, 2019}, pages 3613--3618. Association for Computational Linguistics.

\bibitem[{Chan et~al.(2021)Chan, Ong, Pung, Zhang, and Fu}]{cocon}
Alvin Chan, Yew{-}Soon Ong, Bill Pung, Aston Zhang, and Jie Fu. 2021.
\newblock \href {https://openreview.net/forum?id=VD\_ozqvBy4W} {Cocon: {A} self-supervised approach for controlled text generation}.
\newblock In \emph{9th International Conference on Learning Representations, {ICLR} 2021, Virtual Event, Austria, May 3-7, 2021}. OpenReview.net.

\bibitem[{Chen et~al.(2021)Chen, Lu, Rajeswaran, Lee, Grover, Laskin, Abbeel, Srinivas, and Mordatch}]{decision}
Lili Chen, Kevin Lu, Aravind Rajeswaran, Kimin Lee, Aditya Grover, Michael Laskin, Pieter Abbeel, Aravind Srinivas, and Igor Mordatch. 2021.
\newblock \href {https://proceedings.neurips.cc/paper/2021/hash/7f489f642a0ddb10272b5c31057f0663-Abstract.html} {Decision transformer: Reinforcement learning via sequence modeling}.
\newblock In \emph{Advances in Neural Information Processing Systems 34: Annual Conference on Neural Information Processing Systems 2021, NeurIPS 2021, December 6-14, 2021, virtual}, pages 15084--15097.

\bibitem[{Dathathri et~al.(2020)Dathathri, Madotto, Lan, Hung, Frank, Molino, Yosinski, and Liu}]{pplm}
Sumanth Dathathri, Andrea Madotto, Janice Lan, Jane Hung, Eric Frank, Piero Molino, Jason Yosinski, and Rosanne Liu. 2020.
\newblock \href {https://openreview.net/forum?id=H1edEyBKDS} {Plug and play language models: {A} simple approach to controlled text generation}.
\newblock In \emph{8th International Conference on Learning Representations, {ICLR} 2020, Addis Ababa, Ethiopia, April 26-30, 2020}. OpenReview.net.

\bibitem[{Fan et~al.(2018)Fan, Lewis, and Dauphin}]{writingprompts}
Angela Fan, Mike Lewis, and Yann~N. Dauphin. 2018.
\newblock \href {https://doi.org/10.18653/V1/P18-1082} {Hierarchical neural story generation}.
\newblock In \emph{Proceedings of the 56th Annual Meeting of the Association for Computational Linguistics, {ACL} 2018, Melbourne, Australia, July 15-20, 2018, Volume 1: Long Papers}, pages 889--898. Association for Computational Linguistics.

\bibitem[{Ficler and Goldberg(2017)}]{tense}
Jessica Ficler and Yoav Goldberg. 2017.
\newblock \href {http://arxiv.org/abs/1707.02633} {Controlling linguistic style aspects in neural language generation}.
\newblock \emph{CoRR}, abs/1707.02633.

\bibitem[{Gehman et~al.(2020)Gehman, Gururangan, Sap, Choi, and Smith}]{realtoxic}
Samuel Gehman, Suchin Gururangan, Maarten Sap, Yejin Choi, and Noah~A. Smith. 2020.
\newblock \href {https://doi.org/10.18653/V1/2020.FINDINGS-EMNLP.301} {Realtoxicityprompts: Evaluating neural toxic degeneration in language models}.
\newblock In \emph{Findings of the Association for Computational Linguistics: {EMNLP} 2020, Online Event, 16-20 November 2020}, volume {EMNLP} 2020 of \emph{Findings of {ACL}}, pages 3356--3369. Association for Computational Linguistics.

\bibitem[{Gu et~al.(2022{\natexlab{a}})Gu, Tinn, Cheng, Lucas, Usuyama, Liu, Naumann, Gao, and Poon}]{apply-2}
Yu~Gu, Robert Tinn, Hao Cheng, Michael Lucas, Naoto Usuyama, Xiaodong Liu, Tristan Naumann, Jianfeng Gao, and Hoifung Poon. 2022{\natexlab{a}}.
\newblock \href {https://doi.org/10.1145/3458754} {Domain-specific language model pretraining for biomedical natural language processing}.
\newblock \emph{{ACM} Trans. Comput. Heal.}, 3(1):2:1--2:23.

\bibitem[{Gu et~al.(2022{\natexlab{b}})Gu, Feng, Ma, Zhang, Gong, and Qin}]{distlens}
Yuxuan Gu, Xiaocheng Feng, Sicheng Ma, Lingyuan Zhang, Heng Gong, and Bing Qin. 2022{\natexlab{b}}.
\newblock \href {https://doi.org/10.18653/V1/2022.EMNLP-MAIN.67} {A distributional lens for multi-aspect controllable text generation}.
\newblock In \emph{Proceedings of the 2022 Conference on Empirical Methods in Natural Language Processing, {EMNLP} 2022, Abu Dhabi, United Arab Emirates, December 7-11, 2022}, pages 1023--1043. Association for Computational Linguistics.

\bibitem[{Gu et~al.(2023)Gu, Feng, Ma, Zhang, Gong, Zhong, and Qin}]{normal_flow}
Yuxuan Gu, Xiaocheng Feng, Sicheng Ma, Lingyuan Zhang, Heng Gong, Weihong Zhong, and Bing Qin. 2023.
\newblock \href {https://doi.org/10.18653/v1/2023.acl-long.704} {Controllable text generation via probability density estimation in the latent space}.
\newblock In \emph{Proceedings of the 61st Annual Meeting of the Association for Computational Linguistics (Volume 1: Long Papers), {ACL} 2023, Toronto, Canada, July 9-14, 2023}, pages 12590--12616. Association for Computational Linguistics.

\bibitem[{Gururangan et~al.(2020)Gururangan, Marasovic, Swayamdipta, Lo, Beltagy, Downey, and Smith}]{pretrain}
Suchin Gururangan, Ana Marasovic, Swabha Swayamdipta, Kyle Lo, Iz~Beltagy, Doug Downey, and Noah~A. Smith. 2020.
\newblock \href {https://doi.org/10.18653/v1/2020.acl-main.740} {Don't stop pretraining: Adapt language models to domains and tasks}.
\newblock In \emph{Proceedings of the 58th Annual Meeting of the Association for Computational Linguistics, {ACL} 2020, Online, July 5-10, 2020}, pages 8342--8360. Association for Computational Linguistics.

\bibitem[{Huang et~al.(2023)Huang, Liu, Li, Li, Sun, and Liu}]{prompt_gating}
Xuancheng Huang, Zijun Liu, Peng Li, Tao Li, Maosong Sun, and Yang Liu. 2023.
\newblock \href {https://doi.org/10.18653/v1/2023.acl-long.849} {An extensible plug-and-play method for multi-aspect controllable text generation}.
\newblock In \emph{Proceedings of the 61st Annual Meeting of the Association for Computational Linguistics (Volume 1: Long Papers), {ACL} 2023, Toronto, Canada, July 9-14, 2023}, pages 15233--15256. Association for Computational Linguistics.

\bibitem[{Janner et~al.(2021)Janner, Li, and Levine}]{rl_transformer1}
Michael Janner, Qiyang Li, and Sergey Levine. 2021.
\newblock \href {https://proceedings.neurips.cc/paper/2021/hash/099fe6b0b444c23836c4a5d07346082b-Abstract.html} {Offline reinforcement learning as one big sequence modeling problem}.
\newblock In \emph{Advances in Neural Information Processing Systems 34: Annual Conference on Neural Information Processing Systems 2021, NeurIPS 2021, December 6-14, 2021, virtual}, pages 1273--1286.

\bibitem[{Keskar et~al.(2019)Keskar, McCann, Varshney, Xiong, and Socher}]{ctrl}
Nitish~Shirish Keskar, Bryan McCann, Lav~R. Varshney, Caiming Xiong, and Richard Socher. 2019.
\newblock \href {http://arxiv.org/abs/1909.05858} {{CTRL:} {A} conditional transformer language model for controllable generation}.
\newblock \emph{CoRR}, abs/1909.05858.

\bibitem[{Khalifa et~al.(2021)Khalifa, Elsahar, and Dymetman}]{distributional}
Muhammad Khalifa, Hady Elsahar, and Marc Dymetman. 2021.
\newblock \href {https://openreview.net/forum?id=jWkw45-9AbL} {A distributional approach to controlled text generation}.
\newblock In \emph{9th International Conference on Learning Representations, {ICLR} 2021, Virtual Event, Austria, May 3-7, 2021}. OpenReview.net.

\bibitem[{Krause et~al.(2021)Krause, Gotmare, McCann, Keskar, Joty, Socher, and Rajani}]{gedi}
Ben Krause, Akhilesh~Deepak Gotmare, Bryan McCann, Nitish~Shirish Keskar, Shafiq~R. Joty, Richard Socher, and Nazneen~Fatema Rajani. 2021.
\newblock \href {https://doi.org/10.18653/v1/2021.findings-emnlp.424} {Gedi: Generative discriminator guided sequence generation}.
\newblock In \emph{Findings of the Association for Computational Linguistics: {EMNLP} 2021, Virtual Event / Punta Cana, Dominican Republic, 16-20 November, 2021}, pages 4929--4952. Association for Computational Linguistics.

\bibitem[{Kumar et~al.(2021)Kumar, Malmi, Severyn, and Tsvetkov}]{mucoco}
Sachin Kumar, Eric Malmi, Aliaksei Severyn, and Yulia Tsvetkov. 2021.
\newblock \href {https://proceedings.neurips.cc/paper/2021/hash/79ec2a4246feb2126ecf43c4a4418002-Abstract.html} {Controlled text generation as continuous optimization with multiple constraints}.
\newblock In \emph{Advances in Neural Information Processing Systems 34: Annual Conference on Neural Information Processing Systems 2021, NeurIPS 2021, December 6-14, 2021, virtual}, pages 14542--14554.

\bibitem[{Lample et~al.(2019)Lample, Subramanian, Smith, Denoyer, Ranzato, and Boureau}]{yelp}
Guillaume Lample, Sandeep Subramanian, Eric~Michael Smith, Ludovic Denoyer, Marc'Aurelio Ranzato, and Y{-}Lan Boureau. 2019.
\newblock \href {https://openreview.net/forum?id=H1g2NhC5KQ} {Multiple-attribute text rewriting}.
\newblock In \emph{7th International Conference on Learning Representations, {ICLR} 2019, New Orleans, LA, USA, May 6-9, 2019}. OpenReview.net.

\bibitem[{Li et~al.(2016)Li, Galley, Brockett, Gao, and Dolan}]{dist}
Jiwei Li, Michel Galley, Chris Brockett, Jianfeng Gao, and Bill Dolan. 2016.
\newblock \href {https://doi.org/10.18653/V1/N16-1014} {A diversity-promoting objective function for neural conversation models}.
\newblock In \emph{{NAACL} {HLT} 2016, The 2016 Conference of the North American Chapter of the Association for Computational Linguistics: Human Language Technologies, San Diego California, USA, June 12-17, 2016}, pages 110--119. The Association for Computational Linguistics.

\bibitem[{Li et~al.(2022)Li, Thickstun, Gulrajani, Liang, and Hashimoto}]{diffusionlm}
Xiang Li, John Thickstun, Ishaan Gulrajani, Percy Liang, and Tatsunori~B. Hashimoto. 2022.
\newblock \href {http://papers.nips.cc/paper\_files/paper/2022/hash/1be5bc25d50895ee656b8c2d9eb89d6a-Abstract-Conference.html} {Diffusion-lm improves controllable text generation}.
\newblock In \emph{NeurIPS}.

\bibitem[{Li and Liang(2021)}]{continuous_prompt}
Xiang~Lisa Li and Percy Liang. 2021.
\newblock \href {https://doi.org/10.18653/v1/2021.acl-long.353} {Prefix-tuning: Optimizing continuous prompts for generation}.
\newblock In \emph{Proceedings of the 59th Annual Meeting of the Association for Computational Linguistics and the 11th International Joint Conference on Natural Language Processing, {ACL/IJCNLP} 2021, (Volume 1: Long Papers), Virtual Event, August 1-6, 2021}, pages 4582--4597. Association for Computational Linguistics.

\bibitem[{Lin and Riedl(2021)}]{blend}
Zhiyu Lin and Mark~O. Riedl. 2021.
\newblock \href {https://ojs.aaai.org/index.php/AIIDE/article/view/18891} {Plug-and-blend: {A} framework for plug-and-play controllable story generation with sketches}.
\newblock In \emph{Proceedings of the Seventeenth {AAAI} Conference on Artificial Intelligence and Interactive Digital Entertainment, {AIIDE} 2021, virtual, October 11-15, 2021}, pages 58--65. {AAAI} Press.

\bibitem[{Liu et~al.(2021)Liu, Sap, Lu, Swayamdipta, Bhagavatula, Smith, and Choi}]{dexpert}
Alisa Liu, Maarten Sap, Ximing Lu, Swabha Swayamdipta, Chandra Bhagavatula, Noah~A. Smith, and Yejin Choi. 2021.
\newblock \href {https://doi.org/10.18653/v1/2021.acl-long.522} {Dexperts: Decoding-time controlled text generation with experts and anti-experts}.
\newblock In \emph{Proceedings of the 59th Annual Meeting of the Association for Computational Linguistics and the 11th International Joint Conference on Natural Language Processing, {ACL/IJCNLP} 2021, (Volume 1: Long Papers), Virtual Event, August 1-6, 2021}, pages 6691--6706. Association for Computational Linguistics.

\bibitem[{Liu et~al.(2019)Liu, Ott, Goyal, Du, Joshi, Chen, Levy, Lewis, Zettlemoyer, and Stoyanov}]{roberta}
Yinhan Liu, Myle Ott, Naman Goyal, Jingfei Du, Mandar Joshi, Danqi Chen, Omer Levy, Mike Lewis, Luke Zettlemoyer, and Veselin Stoyanov. 2019.
\newblock \href {http://arxiv.org/abs/1907.11692} {Roberta: {A} robustly optimized {BERT} pretraining approach}.
\newblock \emph{CoRR}, abs/1907.11692.

\bibitem[{Lu et~al.(2022)Lu, Welleck, Hessel, Jiang, Qin, West, Ammanabrolu, and Choi}]{quark}
Ximing Lu, Sean Welleck, Jack Hessel, Liwei Jiang, Lianhui Qin, Peter West, Prithviraj Ammanabrolu, and Yejin Choi. 2022.
\newblock \href {http://papers.nips.cc/paper\_files/paper/2022/hash/b125999bde7e80910cbdbd323087df8f-Abstract-Conference.html} {{QUARK:} controllable text generation with reinforced unlearning}.
\newblock In \emph{NeurIPS}.

\bibitem[{Qian et~al.(2022)Qian, Dong, Shen, Wei, and Chen}]{contrastive}
Jing Qian, Li~Dong, Yelong Shen, Furu Wei, and Weizhu Chen. 2022.
\newblock \href {https://doi.org/10.18653/V1/2022.FINDINGS-ACL.229} {Controllable natural language generation with contrastive prefixes}.
\newblock In \emph{Findings of the Association for Computational Linguistics: {ACL} 2022, Dublin, Ireland, May 22-27, 2022}, pages 2912--2924. Association for Computational Linguistics.

\bibitem[{Rae et~al.(2021)Rae, Borgeaud, Cai, Millican, Hoffmann, Song, Aslanides, Henderson, Ring, Young, Rutherford, Hennigan, Menick, Cassirer, Powell, van~den Driessche, Hendricks, Rauh, Huang, Glaese, Welbl, Dathathri, Huang, Uesato, Mellor, Higgins, Creswell, McAleese, Wu, Elsen, Jayakumar, Buchatskaya, Budden, Sutherland, Simonyan, Paganini, Sifre, Martens, Li, Kuncoro, Nematzadeh, Gribovskaya, Donato, Lazaridou, Mensch, Lespiau, Tsimpoukelli, Grigorev, Fritz, Sottiaux, Pajarskas, Pohlen, Gong, Toyama, de~Masson~d'Autume, Li, Terzi, Mikulik, Babuschkin, Clark, de~Las~Casas, Guy, Jones, Bradbury, Johnson, Hechtman, Weidinger, Gabriel, Isaac, Lockhart, Osindero, Rimell, Dyer, Vinyals, Ayoub, Stanway, Bennett, Hassabis, Kavukcuoglu, and Irving}]{toxic-1}
Jack~W. Rae, Sebastian Borgeaud, Trevor Cai, Katie Millican, Jordan Hoffmann, H.~Francis Song, John Aslanides, Sarah Henderson, Roman Ring, Susannah Young, Eliza Rutherford, Tom Hennigan, Jacob Menick, Albin Cassirer, Richard Powell, George van~den Driessche, Lisa~Anne Hendricks, Maribeth Rauh, Po{-}Sen Huang, Amelia Glaese, Johannes Welbl, Sumanth Dathathri, Saffron Huang, Jonathan Uesato, John Mellor, Irina Higgins, Antonia Creswell, Nat McAleese, Amy Wu, Erich Elsen, Siddhant~M. Jayakumar, Elena Buchatskaya, David Budden, Esme Sutherland, Karen Simonyan, Michela Paganini, Laurent Sifre, Lena Martens, Xiang~Lorraine Li, Adhiguna Kuncoro, Aida Nematzadeh, Elena Gribovskaya, Domenic Donato, Angeliki Lazaridou, Arthur Mensch, Jean{-}Baptiste Lespiau, Maria Tsimpoukelli, Nikolai Grigorev, Doug Fritz, Thibault Sottiaux, Mantas Pajarskas, Toby Pohlen, Zhitao Gong, Daniel Toyama, Cyprien de~Masson~d'Autume, Yujia Li, Tayfun Terzi, Vladimir Mikulik, Igor Babuschkin, Aidan Clark, Diego de~Las~Casas, Aurelia Guy,
  Chris Jones, James Bradbury, Matthew~J. Johnson, Blake~A. Hechtman, Laura Weidinger, Iason Gabriel, William Isaac, Edward Lockhart, Simon Osindero, Laura Rimell, Chris Dyer, Oriol Vinyals, Kareem Ayoub, Jeff Stanway, Lorrayne Bennett, Demis Hassabis, Koray Kavukcuoglu, and Geoffrey Irving. 2021.
\newblock \href {http://arxiv.org/abs/2112.11446} {Scaling language models: Methods, analysis {\&} insights from training gopher}.
\newblock \emph{CoRR}, abs/2112.11446.

\bibitem[{Schulman et~al.(2017)Schulman, Wolski, Dhariwal, Radford, and Klimov}]{PPO}
John Schulman, Filip Wolski, Prafulla Dhariwal, Alec Radford, and Oleg Klimov. 2017.
\newblock \href {http://arxiv.org/abs/1707.06347} {Proximal policy optimization algorithms}.
\newblock \emph{CoRR}, abs/1707.06347.

\bibitem[{Weidinger et~al.(2021)Weidinger, Mellor, Rauh, Griffin, Uesato, Huang, Cheng, Glaese, Balle, Kasirzadeh, Kenton, Brown, Hawkins, Stepleton, Biles, Birhane, Haas, Rimell, Hendricks, Isaac, Legassick, Irving, and Gabriel}]{toxic-2}
Laura Weidinger, John Mellor, Maribeth Rauh, Conor Griffin, Jonathan Uesato, Po{-}Sen Huang, Myra Cheng, Mia Glaese, Borja Balle, Atoosa Kasirzadeh, Zac Kenton, Sasha Brown, Will Hawkins, Tom Stepleton, Courtney Biles, Abeba Birhane, Julia Haas, Laura Rimell, Lisa~Anne Hendricks, William Isaac, Sean Legassick, Geoffrey Irving, and Iason Gabriel. 2021.
\newblock \href {http://arxiv.org/abs/2112.04359} {Ethical and social risks of harm from language models}.
\newblock \emph{CoRR}, abs/2112.04359.

\bibitem[{Wu et~al.(2023)Wu, Hu, Shi, Dziri, Suhr, Ammanabrolu, Smith, Ostendorf, and Hajishirzi}]{fine-grained-rlhf}
Zeqiu Wu, Yushi Hu, Weijia Shi, Nouha Dziri, Alane Suhr, Prithviraj Ammanabrolu, Noah~A. Smith, Mari Ostendorf, and Hannaneh Hajishirzi. 2023.
\newblock \href {https://doi.org/10.48550/ARXIV.2306.01693} {Fine-grained human feedback gives better rewards for language model training}.
\newblock \emph{CoRR}, abs/2306.01693.

\bibitem[{Xu et~al.(2023)Xu, Lu, Shen, Zhang, Zhao, and Gan}]{rl_transformer3}
Mengdi Xu, Yuchen Lu, Yikang Shen, Shun Zhang, Ding Zhao, and Chuang Gan. 2023.
\newblock \href {https://openreview.net/pdf?id=AatUEvC-Wjv} {Hyper-decision transformer for efficient online policy adaptation}.
\newblock In \emph{The Eleventh International Conference on Learning Representations, {ICLR} 2023, Kigali, Rwanda, May 1-5, 2023}. OpenReview.net.

\bibitem[{Yang and Klein(2021)}]{fudge}
Kevin Yang and Dan Klein. 2021.
\newblock \href {https://doi.org/10.18653/v1/2021.naacl-main.276} {{FUDGE:} controlled text generation with future discriminators}.
\newblock In \emph{Proceedings of the 2021 Conference of the North American Chapter of the Association for Computational Linguistics: Human Language Technologies, {NAACL-HLT} 2021, Online, June 6-11, 2021}, pages 3511--3535. Association for Computational Linguistics.

\bibitem[{Yang et~al.(2023{\natexlab{a}})Yang, Liu, Lei, Yang, Xue, Chen, and Xie}]{tailor}
Kexin Yang, Dayiheng Liu, Wenqiang Lei, Baosong Yang, Mingfeng Xue, Boxing Chen, and Jun Xie. 2023{\natexlab{a}}.
\newblock \href {https://doi.org/10.18653/v1/2023.acl-long.25} {Tailor: {A} soft-prompt-based approach to attribute-based controlled text generation}.
\newblock In \emph{Proceedings of the 61st Annual Meeting of the Association for Computational Linguistics (Volume 1: Long Papers), {ACL} 2023, Toronto, Canada, July 9-14, 2023}, pages 410--427. Association for Computational Linguistics.

\bibitem[{Yang et~al.(2023{\natexlab{b}})Yang, Zhang, Xia, Feng, Xiong, and Zhou}]{token-level-rlhf}
Shentao Yang, Shujian Zhang, Congying Xia, Yihao Feng, Caiming Xiong, and Mingyuan Zhou. 2023{\natexlab{b}}.
\newblock \href {https://doi.org/10.48550/ARXIV.2306.00398} {Preference-grounded token-level guidance for language model fine-tuning}.
\newblock \emph{CoRR}, abs/2306.00398.

\bibitem[{Zhang and Song(2022)}]{discup}
Hanqing Zhang and Dawei Song. 2022.
\newblock \href {https://aclanthology.org/2022.emnlp-main.223} {Discup: Discriminator cooperative unlikelihood prompt-tuning for controllable text generation}.
\newblock In \emph{Proceedings of the 2022 Conference on Empirical Methods in Natural Language Processing, {EMNLP} 2022, Abu Dhabi, United Arab Emirates, December 7-11, 2022}, pages 3392--3406. Association for Computational Linguistics.

\bibitem[{Zheng et~al.(2022)Zheng, Zhang, and Grover}]{rl_transformer2}
Qinqing Zheng, Amy Zhang, and Aditya Grover. 2022.
\newblock \href {https://proceedings.mlr.press/v162/zheng22c.html} {Online decision transformer}.
\newblock In \emph{International Conference on Machine Learning, {ICML} 2022, 17-23 July 2022, Baltimore, Maryland, {USA}}, volume 162 of \emph{Proceedings of Machine Learning Research}, pages 27042--27059. {PMLR}.

\end{thebibliography}
